\documentclass[10pt,twocolumn,letterpaper]{article}

\usepackage{times}
\usepackage{epsfig}
\usepackage{graphicx}
\usepackage{amsmath}
\usepackage{amssymb}
\usepackage{epstopdf}
\usepackage{dblfloatfix}
\usepackage{indentfirst}
\usepackage{mathptmx}
\usepackage{sectsty}

\usepackage{caption}
\usepackage{subcaption}

\usepackage[pagebackref=true,breaklinks=true,letterpaper=true,colorlinks,bookmarks=false]{hyperref}

\newsavebox\CBox
\def\textBF#1{\sbox\CBox{#1}\resizebox{\wd\CBox}{\ht\CBox}{\textbf{#1}}}

\newcommand{\etc}{{\emph{etc}}}

\usepackage[top = 2.4cm, bottom = 2.8cm, left = 1.8cm, right = 2.3cm]{geometry}

\sectionfont{\fontsize{12}{15}\selectfont}

\begin{document}

\title{\Large{\bf{Replacing Mobile Camera ISP with a Single Deep Learning Model}}\vspace{2mm}}

\author{Andrey Ignatov\\
{\tt\small andrey@vision.ee.ethz.ch}\\
\and
Luc Van Gool\\
{\tt\small vangool@vision.ee.ethz.ch}\\
\\
ETH Zurich, Switzerland\\
\vspace{-2.8mm}
\and
Radu Timofte\\
{\tt\small timofter@vision.ee.ethz.ch}\\
}
\date{}
\maketitle
\thispagestyle{empty}

\begin{abstract}

\noindent\textit{As the popularity of mobile photography is growing constantly, lots of efforts are being invested now into building complex hand-crafted camera ISP solutions. In this work, we demonstrate that even the most sophisticated ISP pipelines can be replaced with a single end-to-end deep learning model trained without any prior knowledge about the sensor and optics used in a particular device. For this, we present PyNET, a novel pyramidal CNN architecture designed for fine-grained image restoration that implicitly learns to perform all ISP steps such as image demosaicing, denoising, white balancing, color and contrast correction, demoireing, \etc. The model is trained to convert RAW Bayer data obtained directly from mobile camera sensor into photos captured with a professional high-end DSLR camera, making the solution independent of any particular mobile ISP implementation. To validate the proposed approach on the real data, we collected a large-scale dataset consisting of 10 thousand full-resolution RAW--RGB image pairs captured in the wild with the Huawei P20 cameraphone (12.3 MP Sony Exmor IMX380 sensor) and Canon 5D Mark IV DSLR. The experiments demonstrate that the proposed solution can easily get to the level of the embedded P20's ISP pipeline that, unlike our approach, is combining the data from two (RGB + B/W) camera sensors. The dataset, pre-trained models and codes used in this paper are available on the project website~\footnote{\url{http://people.ee.ethz.ch/~ihnatova/pynet.html}}}.
\end{abstract}

\section{Introduction}

While the first mass-market phones and PDAs with mobile cameras appeared in the early 2000s, at the beginning they were producing photos of very low quality, significantly falling behind even the simplest compact cameras. The resolution and quality of mobile photos have been growing constantly since that time, with a substantial boost after 2010, when mobile devices started to get powerful hardware suitable for heavy image signal processing (ISP) systems. Since then, the gap between the quality of photos from smartphones and dedicated point-and-shoot cameras is diminishing rapidly, and the latter ones have become nearly extinct over the past years. With this, smartphones became the main source of photos nowadays, and the role and requirements to their cameras have increased even more.

\begin{figure}[t!]
\centering
\includegraphics[width=0.85\linewidth]{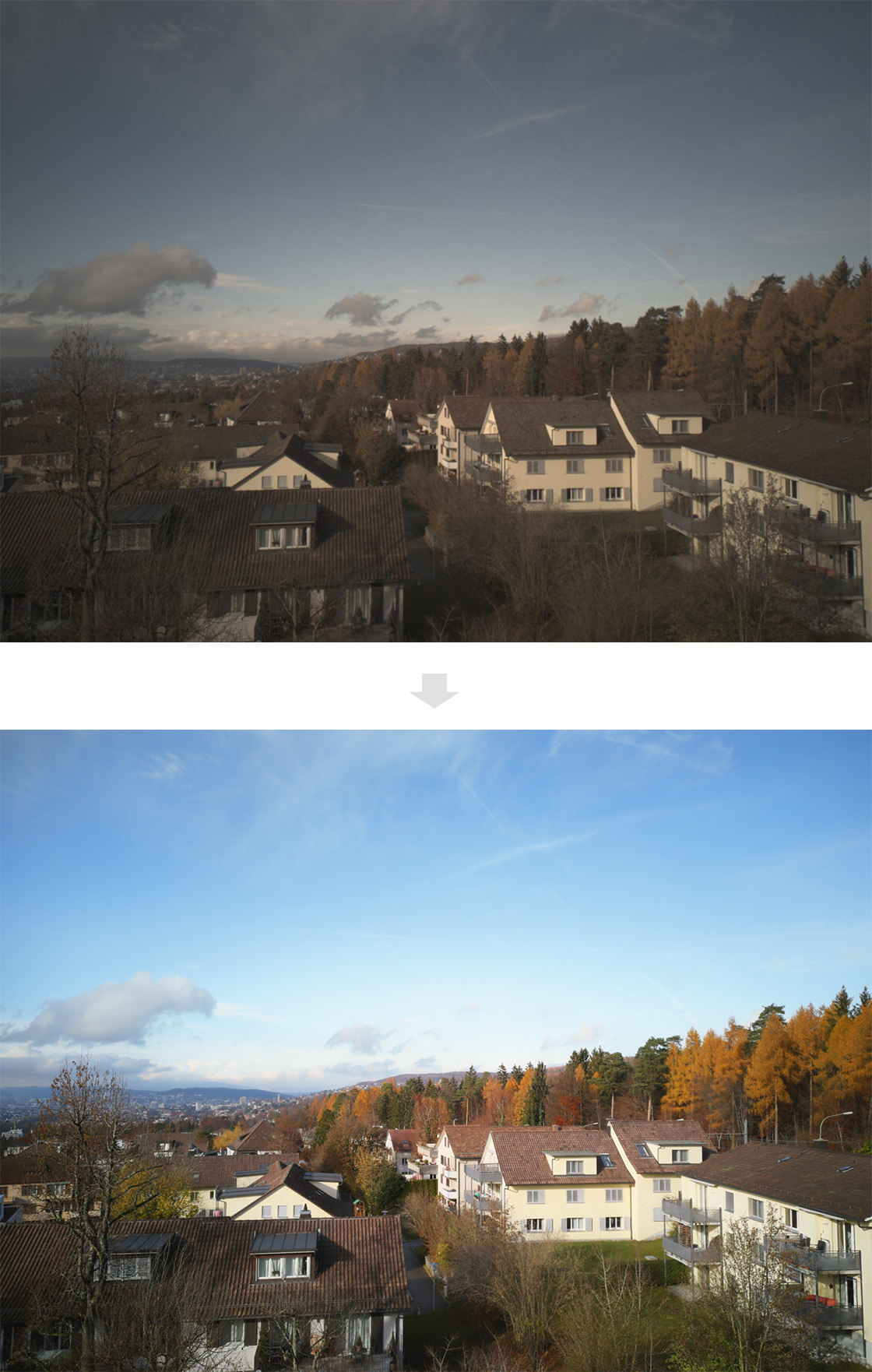}
\vspace{2.4mm}
\caption{\small{Huawei P20 RAW photo (visualized) and the corresponding image reconstructed with our method.}}
\label{fig:Examples}
\vspace{-2.4mm}
\end{figure}

\begin{figure*}[t!]
\centering
\resizebox{1.0\linewidth}{!}
{
\includegraphics[width=0.16\linewidth]{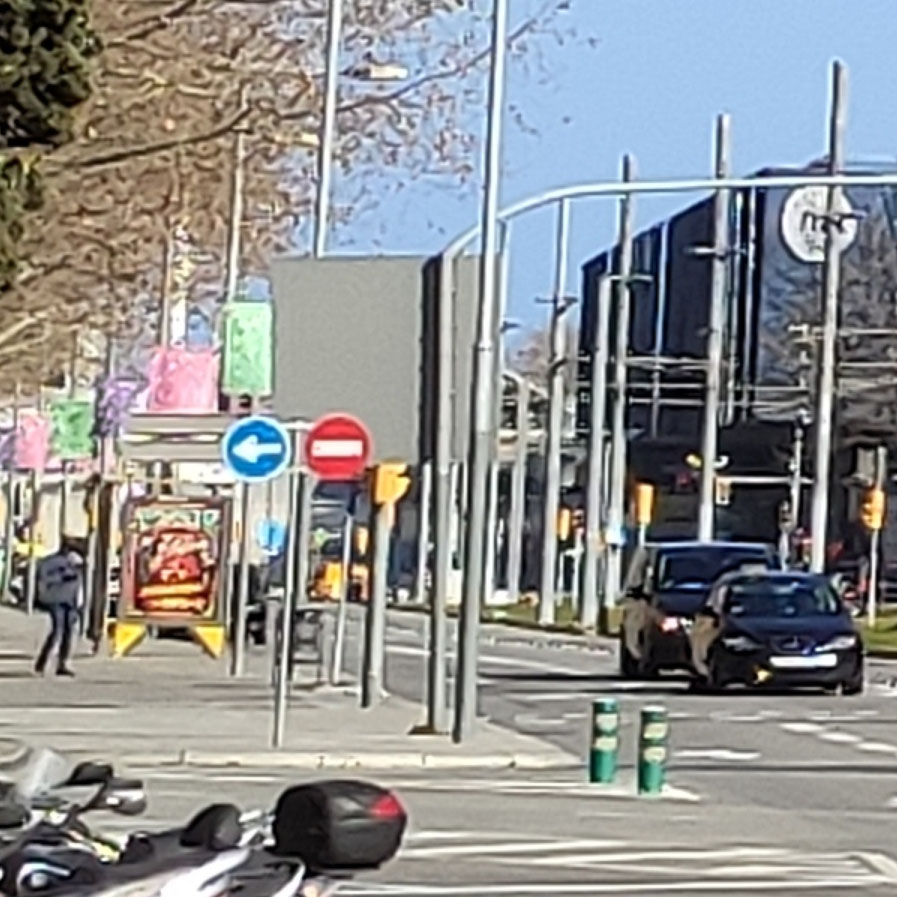}
\includegraphics[width=0.16\linewidth]{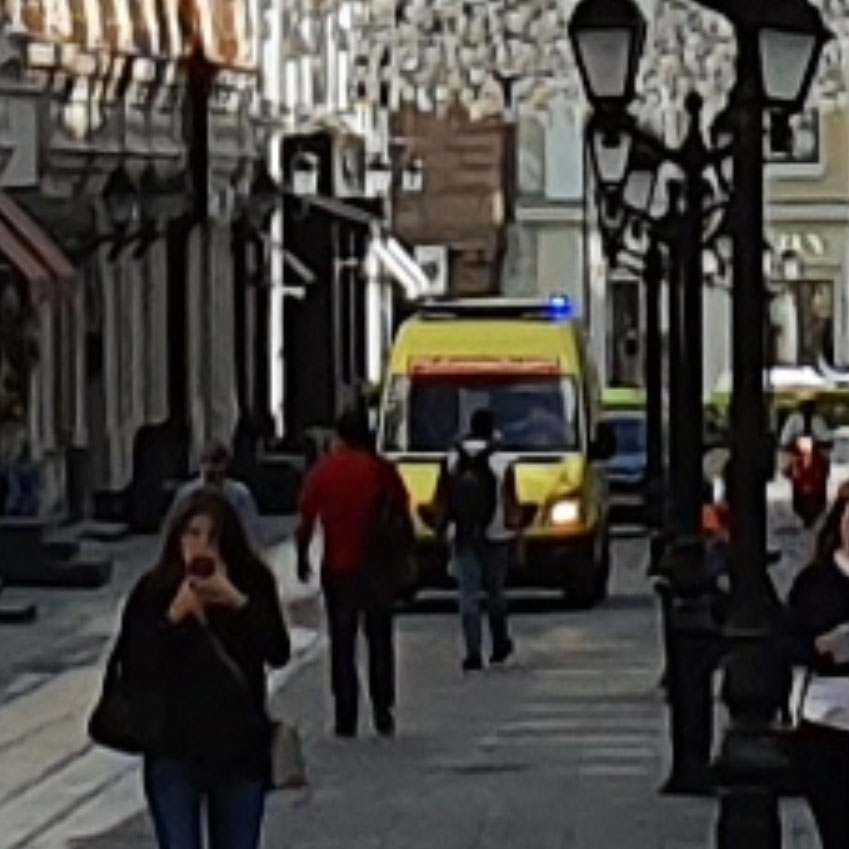}
\includegraphics[width=0.16\linewidth]{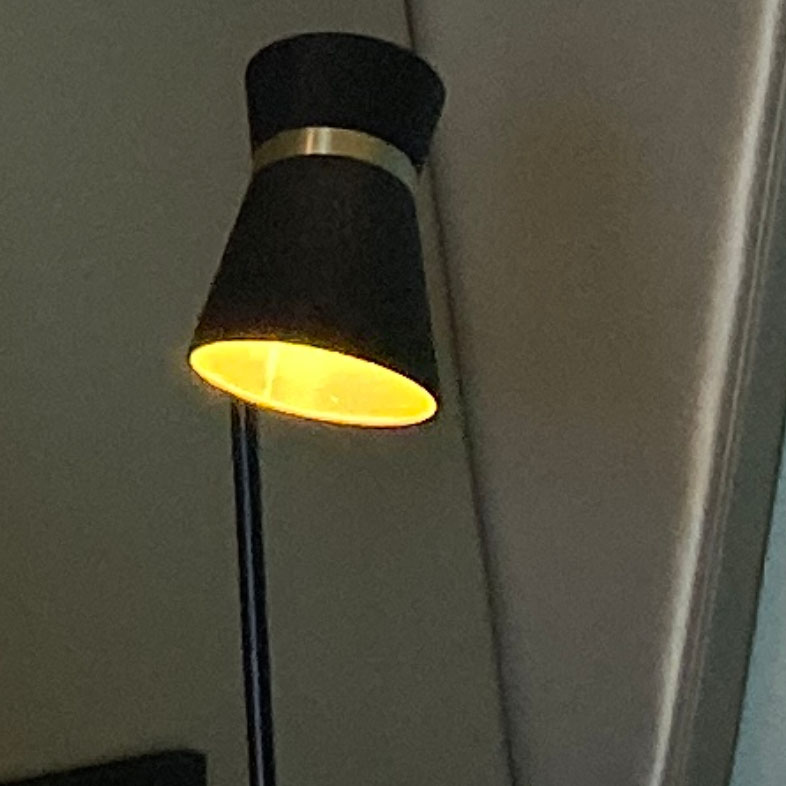}
\includegraphics[width=0.16\linewidth]{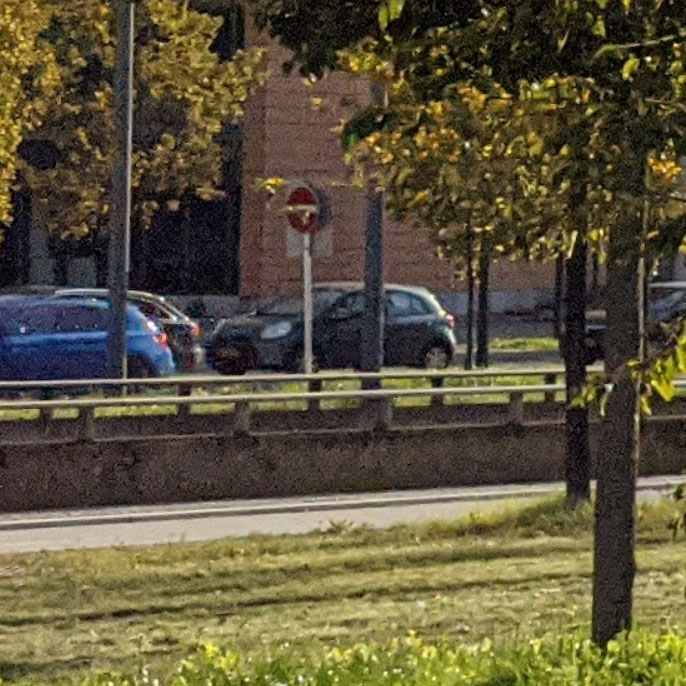}
\includegraphics[width=0.16\linewidth]{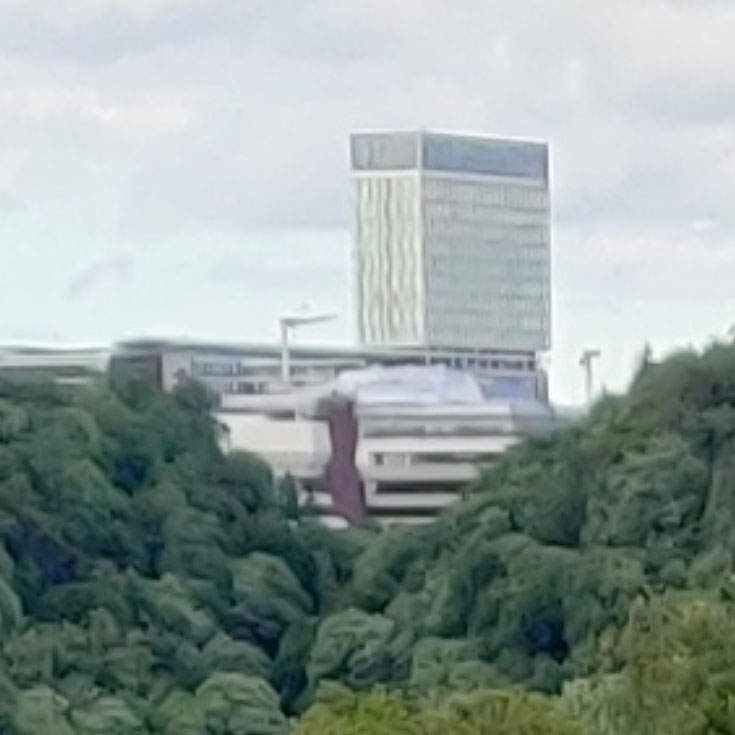}
\includegraphics[width=0.16\linewidth]{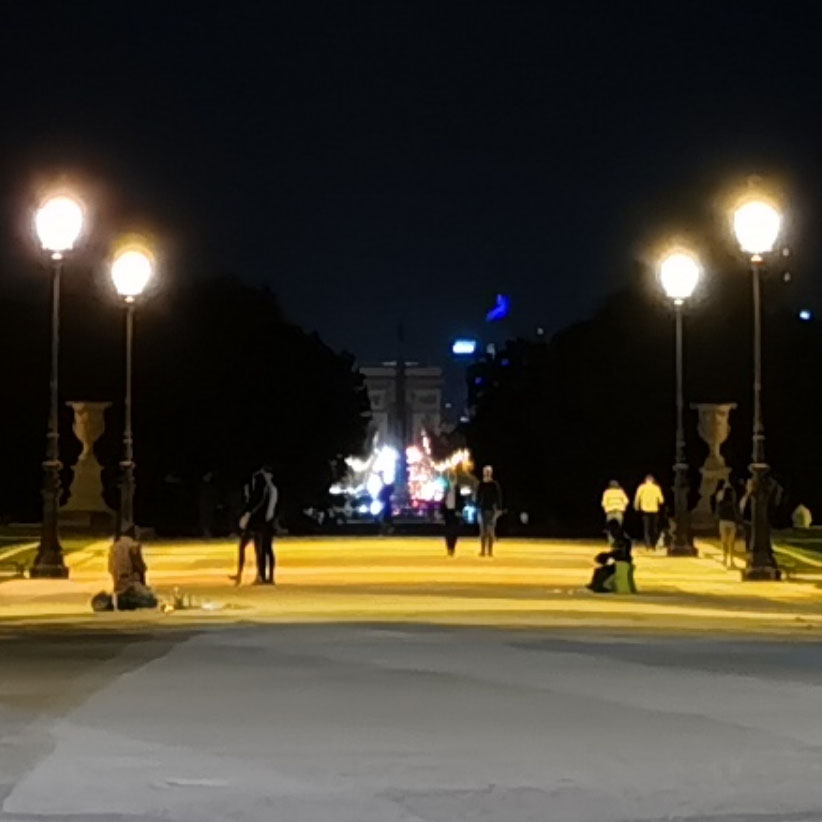}
}
\vspace{-2mm}
\caption{\small{Typical artifacts appearing on photos from mobile cameras. From left to right: cartoonish blurring / ``watercolor effect'' (Xiaomi Mi 9, Samsung Galaxy Note10+), noise (iPhone 11 Pro, Google Pixel 4 XL) and image flattening (OnePlus 7 Pro, Huawei Mate 30 Pro).}}
\label{fig:Artifacts}
\vspace{-1.4mm}
\end{figure*}

The modern mobile ISPs are quite complex software systems that are sequentially solving a number of low-level and global image processing tasks such as image demosaicing, white balance and exposure correction, denoising and sharpening, color and gamma correction, \etc. The parts of the system responsible for different subtasks are usually designed separately, taking into account the particularities of the corresponding sensor and optical system. Despite all the advances in the software stack, the hardware limitations of mobile cameras remain unchanged: small sensors and relatively compact lenses are causing the loss of details, high noise levels and mediocre color rendering. The current classical ISP systems are still unable to handle these issues completely, and are therefore trying to hide them either by flattening the resulting photos or by applying the ``watercolor effect'' that can be found on photos from many recent flagship devices (see Figure~\ref{fig:Artifacts}). Though deep learning models can potentially deal with these problems, and besides that can be also deployed on smartphones having dedicated NPUs and AI chips~\cite{ignatov2019ai,ignatov2018ai}, their current use in mobile ISPs is still limited to scene classification or light photo post-processing.

Unlike the classical approaches, in this paper we propose to learn the entire ISP pipeline with only one deep learning model. For this, we present an architecture that is trained to map RAW Bayer data from the camera sensor to the target high-quality RGB image, thus intrinsically incorporating all image manipulation steps needed for fine-grained photo restoration. Since none of the existing mobile ISPs can produce the required high-quality photos, we are collecting the target RGB images with a professional Canon 5D Mark IV DSLR camera producing clear noise-free high-resolution pictures, and present a large-scale image dataset consisting of 10 thousand RAW (phone) / RGB (DSLR) photo pairs. As for mobile camera, we chose the Huawei P20 cameraphone featuring one of the most sophisticated mobile ISP systems at the time of the dataset collection.

\bigskip

\textBF{Our main contributions are:}

\vspace{-0.8mm}
\begin{itemize}
\setlength\itemsep{-0.2mm}
\item An end-to-end deep learning solution for RAW-to-RGB image mapping problem that is incorporating all image signal processing steps by design.
\item A novel PyNET CNN architecture designed to combine heavy global manipulations with low-level fine-grained image restoration.
\item A large-scale dataset containing 10K RAW--RGB image pairs collected in the wild with the Huawei P20 smartphone and Canon 5D Mark IV DSLR camera.
\item A comprehensive set of experiments evaluating the quantitative and perceptual quality of the reconstructed images, as well as comparing the results of the proposed deep learning approach with the results obtained with the built-in Huawei P20's ISP pipeline.
\end{itemize}
\vspace{-0.8mm}

\section{Related Work}

While the problem of real-world RAW-to-RGB image mapping has not been addressed in the literature, a large number of works dealing with various image restoration and enhancement tasks were proposed during the past years.

Image super-resolution is one of the most classical image reconstruction problems, where the goal is to increase image resolution and sharpness. A large number of efficient solutions were proposed to deal with this task~\cite{agustsson2017ntire,timofte2018ntire}, starting from the simplest CNN approaches~\cite{dong2016image,kim2016accurate,shi2016real} to complex GAN-based systems~\cite{ledig2017photo,sajjadi2017enhancenet,wang2018esrgan}, deep residual models~\cite{lim2017enhanced,zhang2018residual,tong2017image}, laplacian pyramid~\cite{lai2017deep} and channel attention~\cite{zhang2018image} networks. Image deblurring~\cite{chakrabarti2016neural,schuler2015learning,pan2016blind,sun2015learning} and denoising~\cite{zhang2017learning,zhang2017beyond,zhang2016fast,svoboda2016compression} are the other two related tasks targeted at removing blur and noise from the pictures.

\begin{figure*}[t!]
\centering
\setlength{\tabcolsep}{1pt}
\resizebox{\linewidth}{!}
{
\begin{tabular}{cccc}
\scriptsize{Huawei P20 RAW - Visualized}\normalsize & \scriptsize{Huawei P20 ISP}\normalsize & \scriptsize{Canon 5D Mark IV}\normalsize\\
    \includegraphics[width=0.24\linewidth]{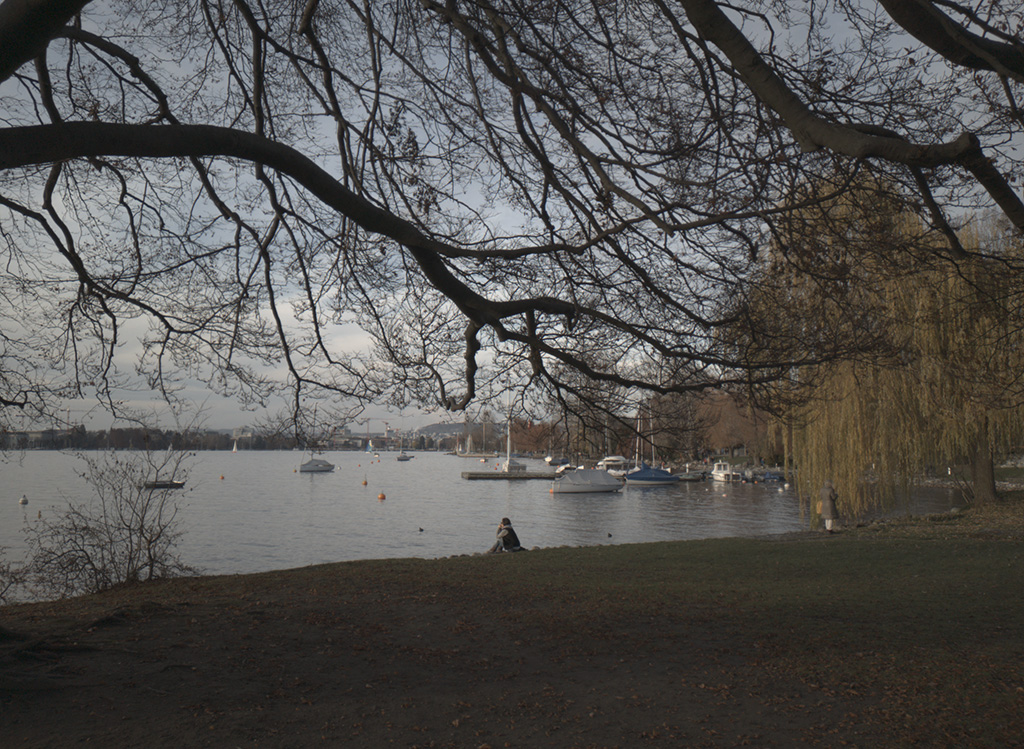}&
    \includegraphics[width=0.24\linewidth]{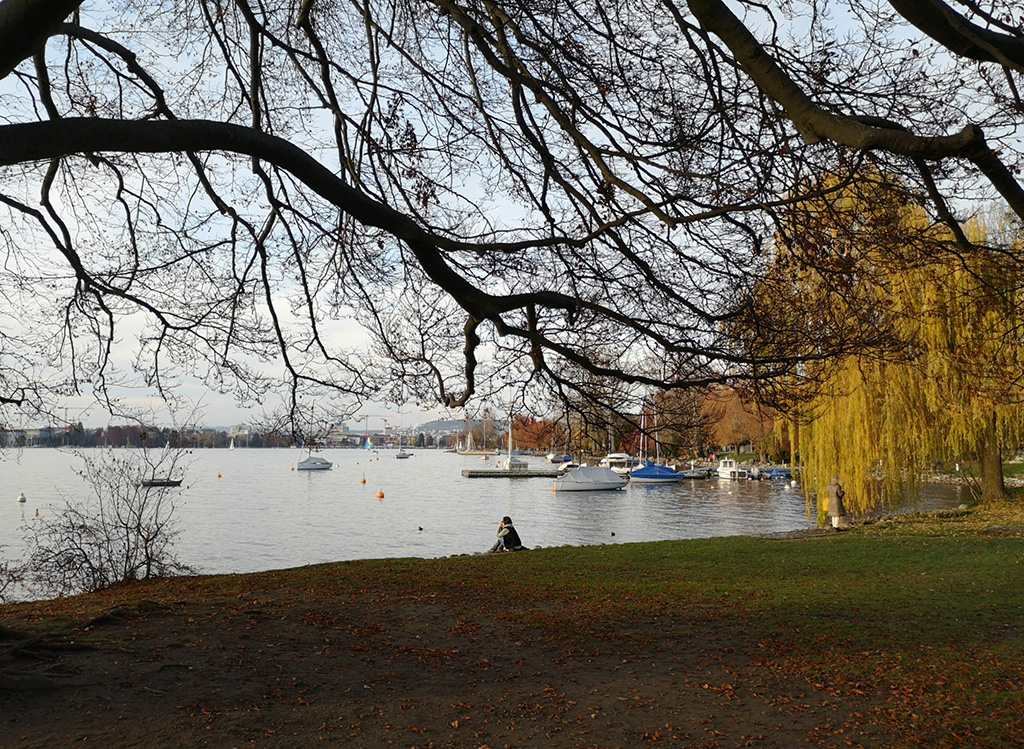}&
    \includegraphics[width=0.24\linewidth]{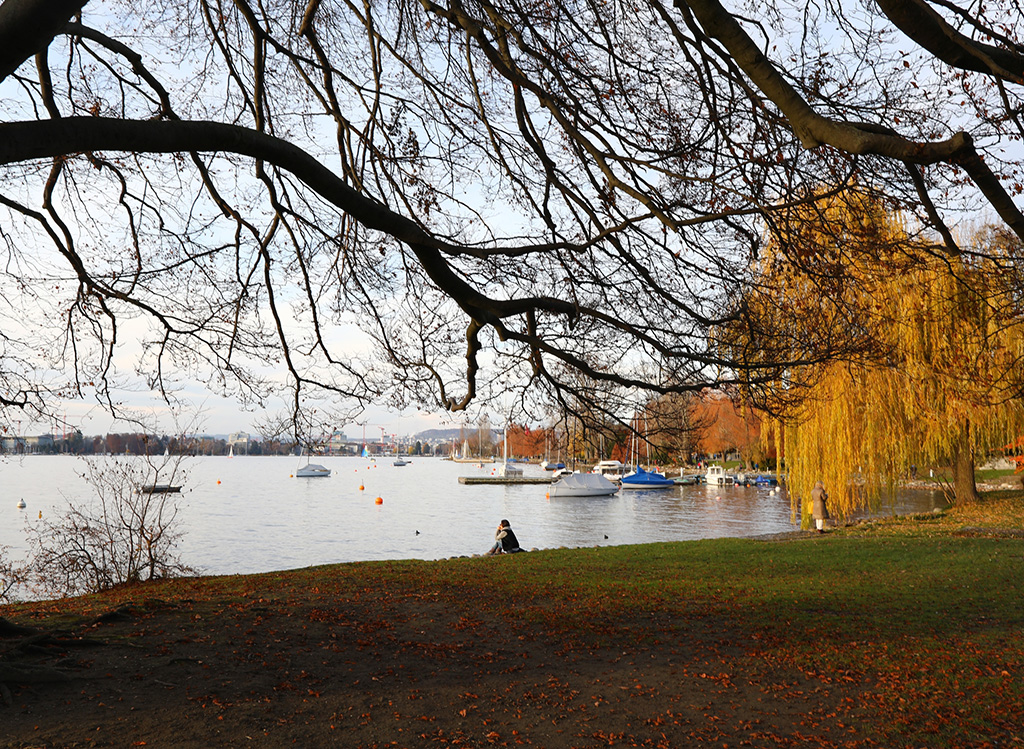}
\end{tabular}
}
\vspace{0.0cm}
\caption{\small{Example set of images from the collected Zurich RAW to RGB dataset. From left to right: original RAW image visualized with a simple ISP script, RGB image obtained with P20's built-in ISP system, and Canon 5D Mark IV target photo.}}
\label{fig:example_photos}
\vspace{-0.2cm}
\end{figure*}

A separate group of tasks encompass various global image adjustment problems. In~\cite{yuan2012automatic,fu2016fusion}, the authors proposed solutions for automatic global luminance and gamma adjustment, while work~\cite{cai2018learning} presented a CNN-based method for image contrast enhancement. In~\cite{yan2016automatic,lee2016automatic}, deep learning solutions for image color and tone corrections were proposed, and in~\cite{salih2012tone,ma2014high} tone mapping algorithms for HDR images were presented.

The problem of comprehensive image quality enhancement was first addressed in~\cite{ignatov2017dslr,ignatov2018wespe}, where the authors proposed to enhance all aspects of low-quality smartphone photos by mapping them to superior-quality images obtained with a high-end reflex camera. The collected DPED dataset was later used in many subsequent works~\cite{zhu2018range,stoutz2018fast,vu2018fast,hui2018perception,liu2018deep} that have significantly improved the results on this problem. Additionally, in~\cite{ignatov2019ntire} the authors examined the possibility of running the resulting image enhancement models directly on smartphones, and proposed a number of efficient solutions for this task. It should be mentioned that though the proposed models were showing nice results, they were targeted at refining the images obtained with smartphone ISPs rather than processing RAW camera data.

While there exist many classical approaches for various image signal processing subtasks such as image demosaicing~\cite{li2008image,dubois2006filter,hirakawa2005adaptive}, denoising~\cite{buades2005non,condat2010simple,foi2008practical}, white balancing~\cite{gijsenij2011improving,van2007edge,buchsbaum1980spatial}, color correction~\cite{kwok2013simultaneous,rizzi2003new,rizzi2004retinex}, \etc, only a few works explored the applicability of deep learning models to these problems. In~\cite{park2019color,syu2018learning}, the authors demonstrated that convolutional neural networks can be used for performing image demosaicing, and outperformed several conventional models in this task. Works~\cite{bianco2015color,hu2017fc4} used CNNs for correcting the white balance of RGB images, and in~\cite{yuan2019aim} deep learning models were applied to synthetic LCDMoire dataset for solving image demoireing problem. In~\cite{baskin2018nice}, the authors collected 110 RAW low-lit images with Samsung S7 phone, and used a CNN model to remove noise and brighten demosaiced RGB images obtained with a simple hand-designed ISP. Finally, in work~\cite{ratnasingam2019deep} RAW images were artificially generated from JPEG photos presented in~\cite{ciurea2003large}, and a CNN was applied to reconstruct the original RGB pictures. In this paper, we will go beyond the constrained artificial settings used in the previous works, and will be solving \textit{all} ISP subtasks on real data simultaneously, trying to outperform the commercial ISP system present in one of the best camera phones released in the past two years.

\section{Zurich RAW to RGB dataset}

To get real data for RAW to RGB mapping problem, a large-scale dataset consisting of 20 thousand photos was collected using Huawei P20 smartphone capturing RAW photos (plus the resulting RGB images obtained with Huawei's built-in ISP), and a professional high-end Canon 5D Mark IV camera with Canon EF 24mm f/1.4L fast lens. RAW data was read from P20's 12.3 MP Sony Exmor IMX380 Bayer camera sensor~-- though this phone has a second 20 MP monochrome camera, it is only used by Huawei's internal ISP system, and the corresponding images cannot be retrieved with any public camera API. The photos were captured in automatic mode, and default settings were used throughout the whole collection procedure. The data was collected over several weeks in a variety of places and in various illumination and weather conditions. An example set of captured images is shown in Figure~\ref{fig:example_photos}.

Since the captured RAW--RGB image pairs are not perfectly aligned, we first performed their matching using the same procedure as in~\cite{ignatov2017dslr}. The images were first aligned globally using SIFT keypoints~\cite{lowe2004distinctive} and RANSAC algorithm~\cite{vedaldi2010vlfeat}. Then, smaller patches of size 448$\times$448 were extracted from the preliminary matched images using a non-overlapping sliding window. Two windows were moving in parallel along the two images from each RAW-RGB pair, and the position of the window on DSLR image was additionally adjusted with small shifts and rotations to maximize the cross-correlation between the observed patches. Patches with cross-correlation less than 0.9 were not included into the dataset to avoid large displacements.
This procedure resulted in 48043 RAW-RGB image pairs (of size 448$\times$448$\times$1 and 448$\times$448$\times$3, respectively) that were later used for training / validation (46.8K) and testing (1.2K) the models. RAW image patches were additionally reshaped into the size of 224$\times$224$\times$4, where the four channels correspond to the four colors of the RGBG Bayer filer.
It should be mentioned that all alignment operations were performed only on RGB DSLR images, therefore RAW photos from Huawei P20 remained unmodified, containing the same values as were obtained from the camera sensor.

\section{Proposed Method}

The problem of RAW to RGB mapping is generally involving both global and local image modifications. The first ones are used to alter the image content and its high-level properties such as brightness, while balance or color rendition, while low-level processing is needed for tasks like texture enhancement, sharpening, noise removal, deblurring, \etc. More importantly, there should be an interaction between global and local modifications, as, for example, content understanding is critical for tasks like texture processing or local color correction. While there exists many deep learning models targeted at one of these two problem types, their application to RAW to RGB mapping or to general image enhancement tasks is leading to the corresponding issues: VGG-~\cite{kim2016accurate}, ResNet-~\cite{ledig2017photo} or DenseNet-based~\cite{huang2017densely} networks cannot alter the image significantly, while models relying on U-Net~\cite{ronneberger2015u} or Pix2Pix~\cite{isola2017image} architectures are not good at improving local image properties. To address this issue, in this paper we propose a novel PyNET CNN architecture that is processing image at different scales and combines the learned global and local features together.

\subsection{PyNET CNN Architecture}

\begin{figure}[t!]
\centering
\includegraphics[width=1.0\linewidth]{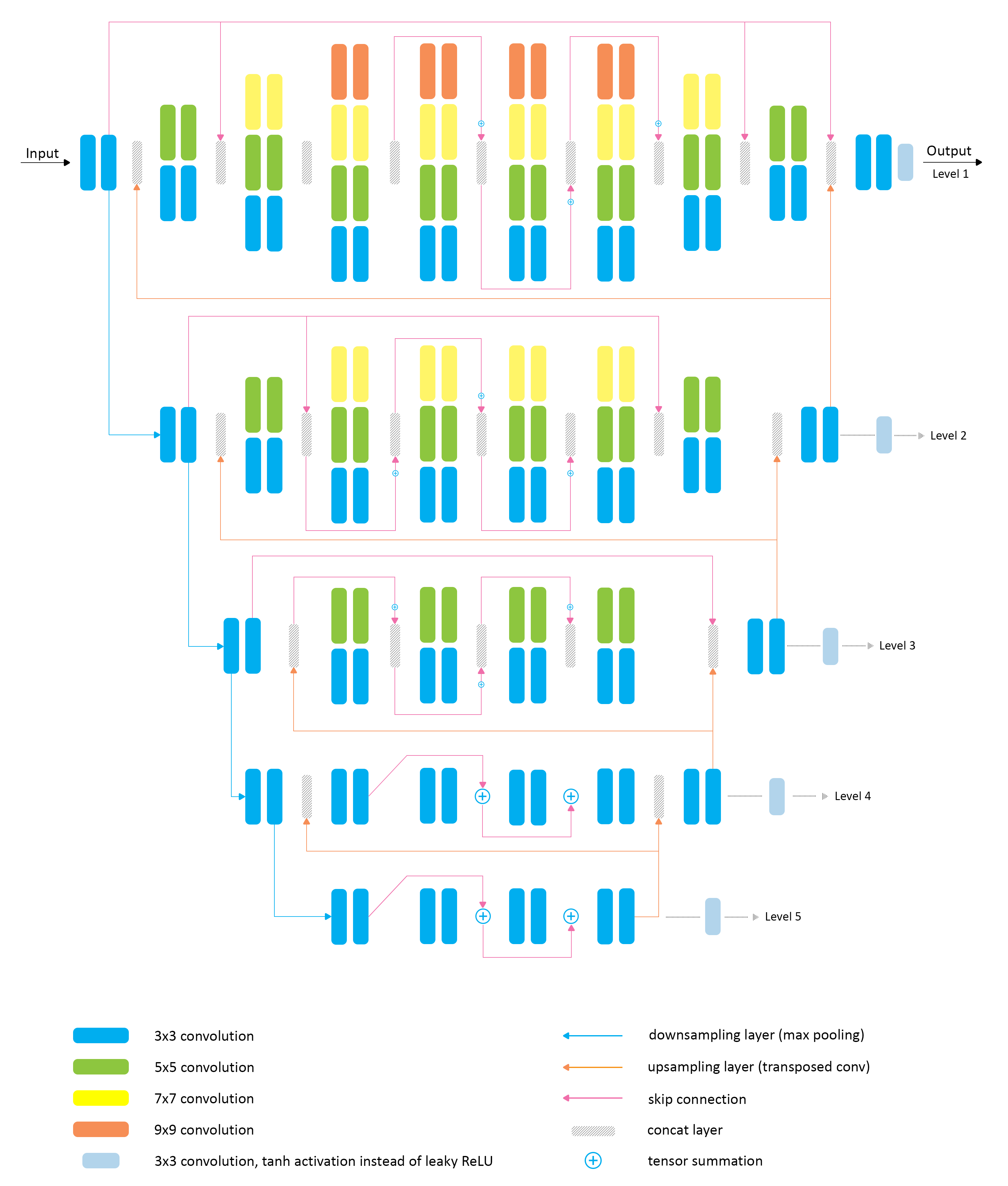}
\vspace{1.4mm}
\caption{\small{The architecture of the proposed PyNET model. Concat and Sum ops are applied to the outputs of the adjacent layers.}}
\label{fig:Architecture}
\vspace{-2.4mm}
\end{figure}

Figure~\ref{fig:Architecture} illustrates schematic representation of the proposed deep learning architecture. The model has an inverted pyramidal shape and is processing the images at five different scales. The proposed architecture has a number of blocks that are processing feature maps in parallel with convolutional filters of different size (from 3$\times$3 to 9$\times$9), and the outputs of the corresponding convolutional layers are then concatenated, which allows the network to learn a more diverse set of features at each level. The outputs obtained at lower scales are upsampled with transposed convolutional layers, stacked with feature maps from the upper level and then subsequently processed in the following convolutional layers. \textit{Leaky ReLU} activation function is applied after each convolutional operation, except for the output layers that are using \textit{tanh} function to map the results to (-1, 1) interval. Instance normalization is used in all convolutional layers that are processing images at lower scales (levels 2-5). We are additionally using one transposed convolutional layer on top of the model that upsamples the images to their target size.

The model is trained sequentially, starting from the lowest layer. This allows to achieve good image reconstruction results at smaller scales that are working with images of very low resolution and performing mostly global image manipulations. After the bottom layer is pre-trained, the same procedure is applied to the next level till the training is done on the original resolution. Since each higher level is getting upscaled high-quality features from the lower part of the model, it mainly learns to reconstruct the missing low-level details and refines the results. Note that the input layer is always the same and is getting images of size 224$\times$224$\times$4, though only a part of the training graph (all layers participating in producing the outputs at the corresponding scale) is trained.

\begin{figure*}[t!]
\centering
\setlength{\tabcolsep}{1pt}
\resizebox{\linewidth}{!}
{
\begin{tabular}{cccc}
\scriptsize{Visualized RAW Image}\normalsize & \scriptsize{Reconstructed RGB Image (PyNET)}\normalsize & \scriptsize{Huawei P20's ISP Photo}\normalsize & \scriptsize{Canon 5D Mark IV Photo}\normalsize\\
    \includegraphics[width=0.24\linewidth]{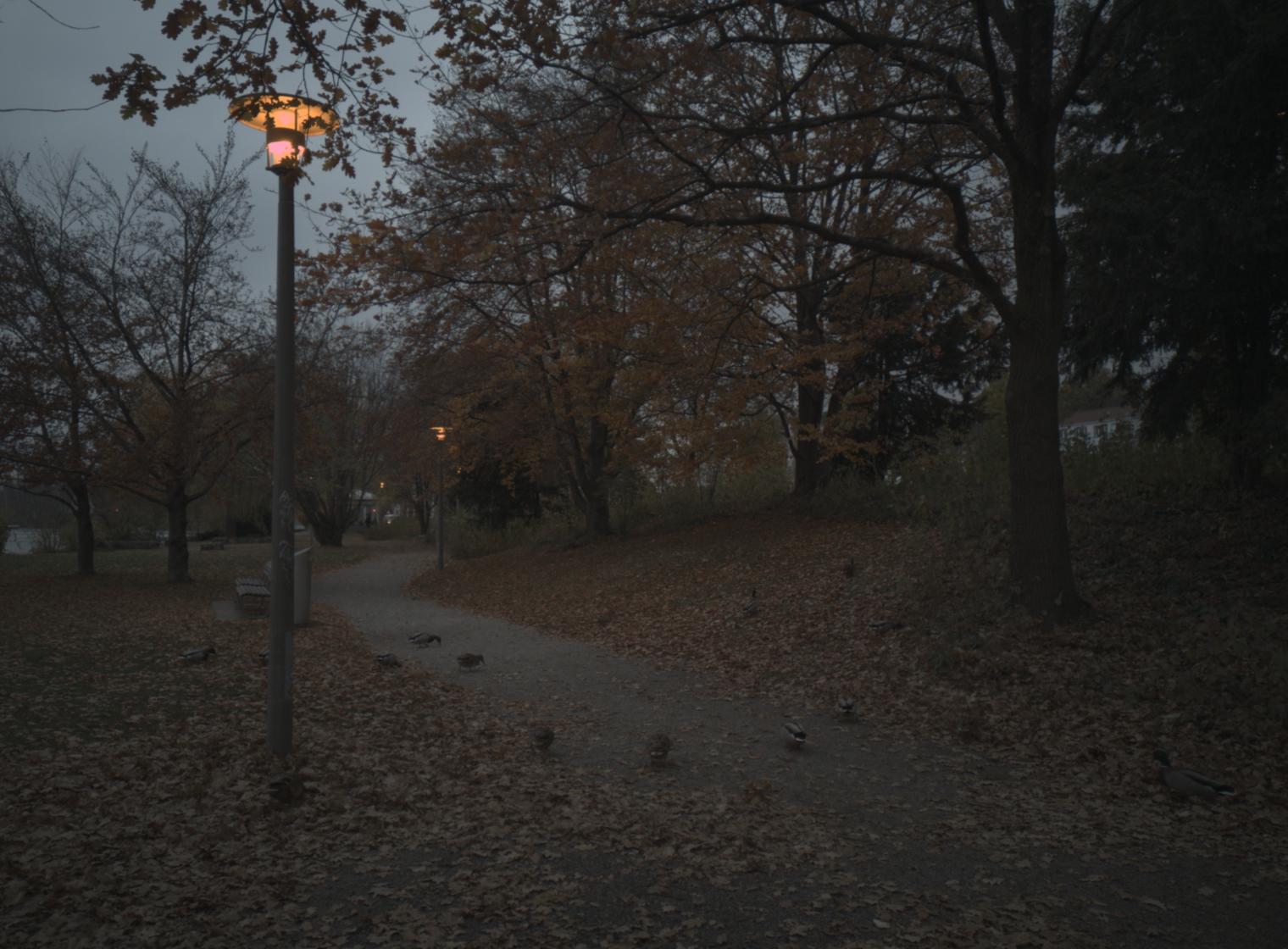}&
    \includegraphics[width=0.24\linewidth]{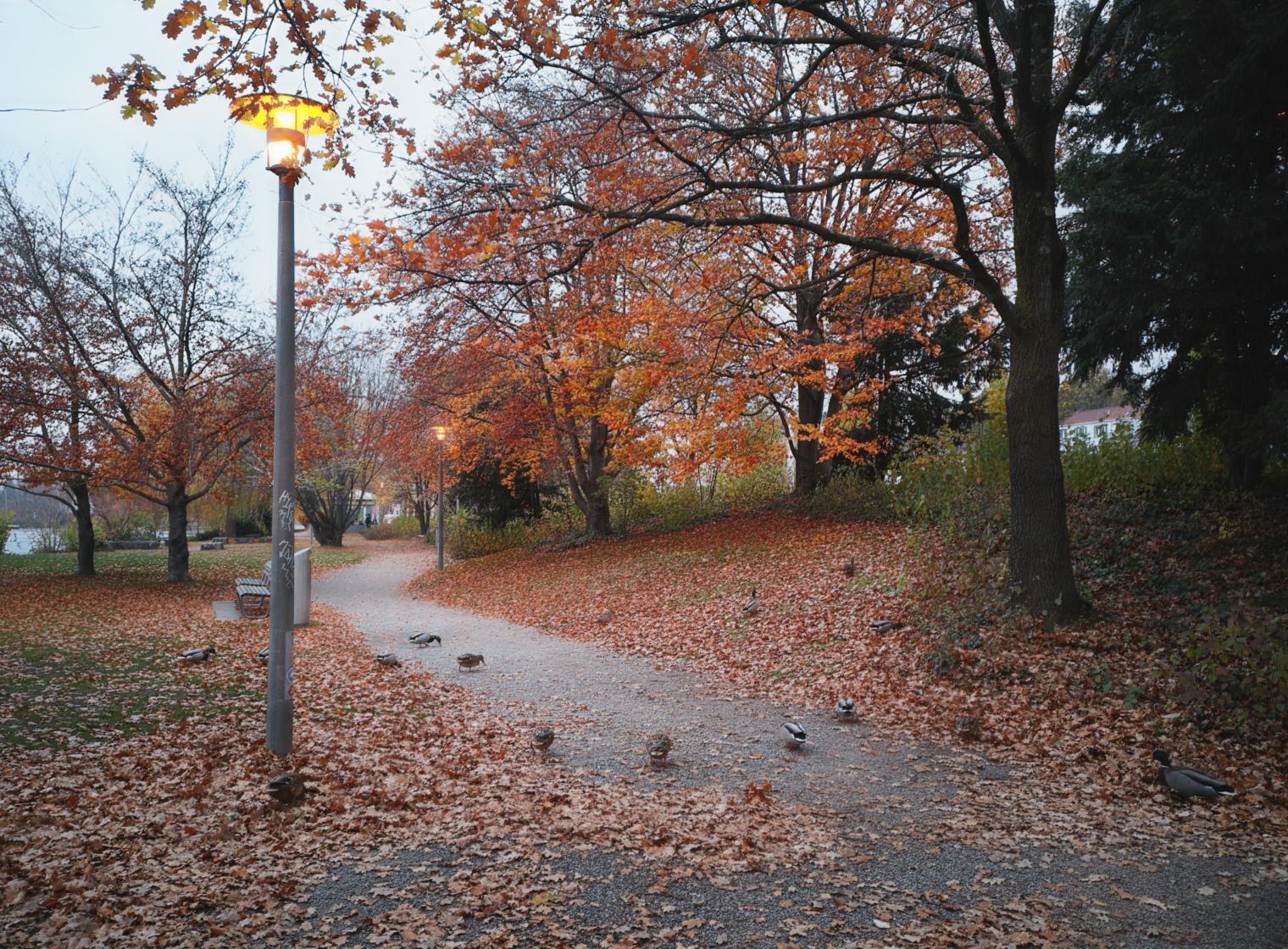}&
    \includegraphics[width=0.24\linewidth]{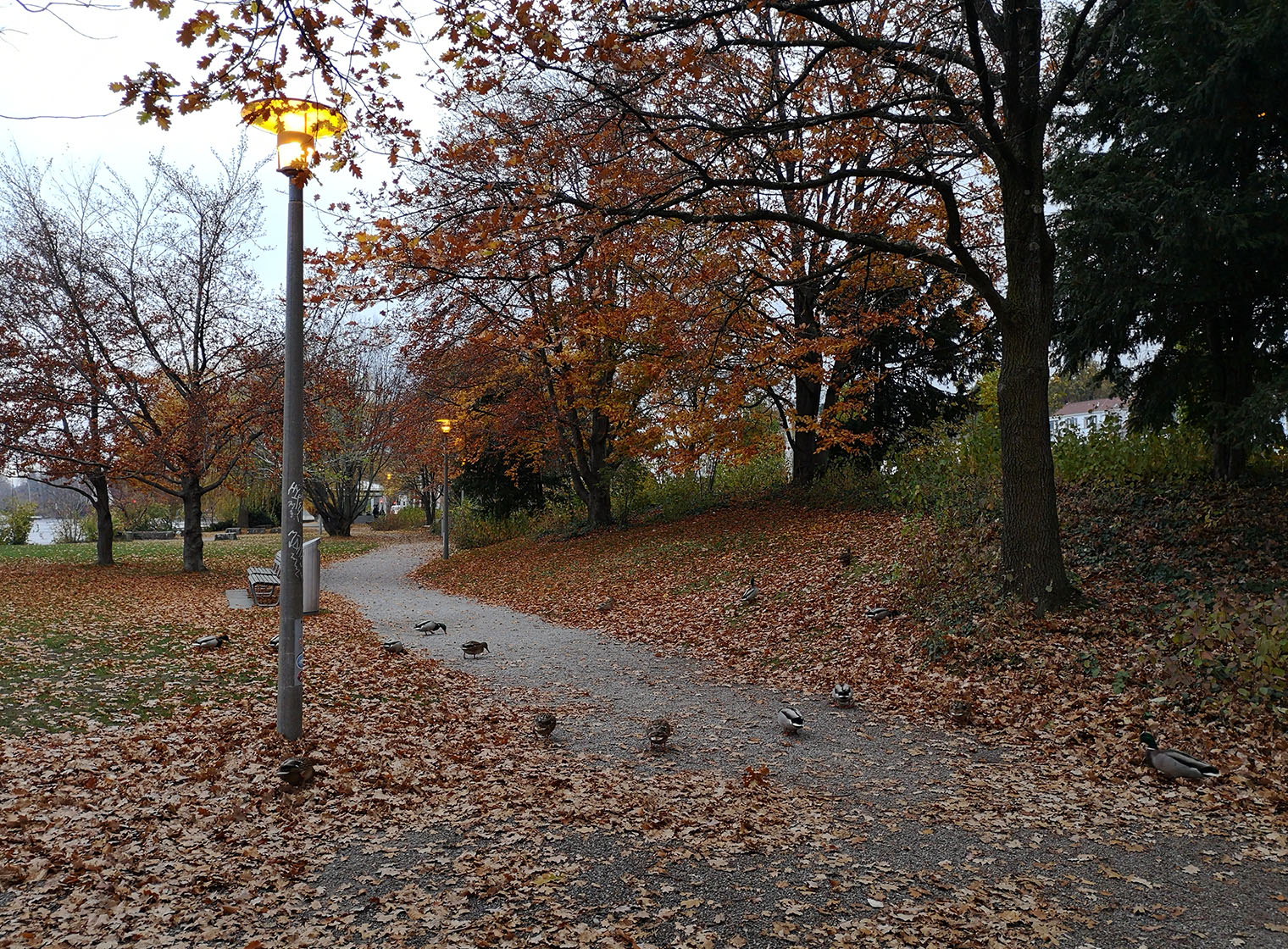}&
    \includegraphics[width=0.24\linewidth]{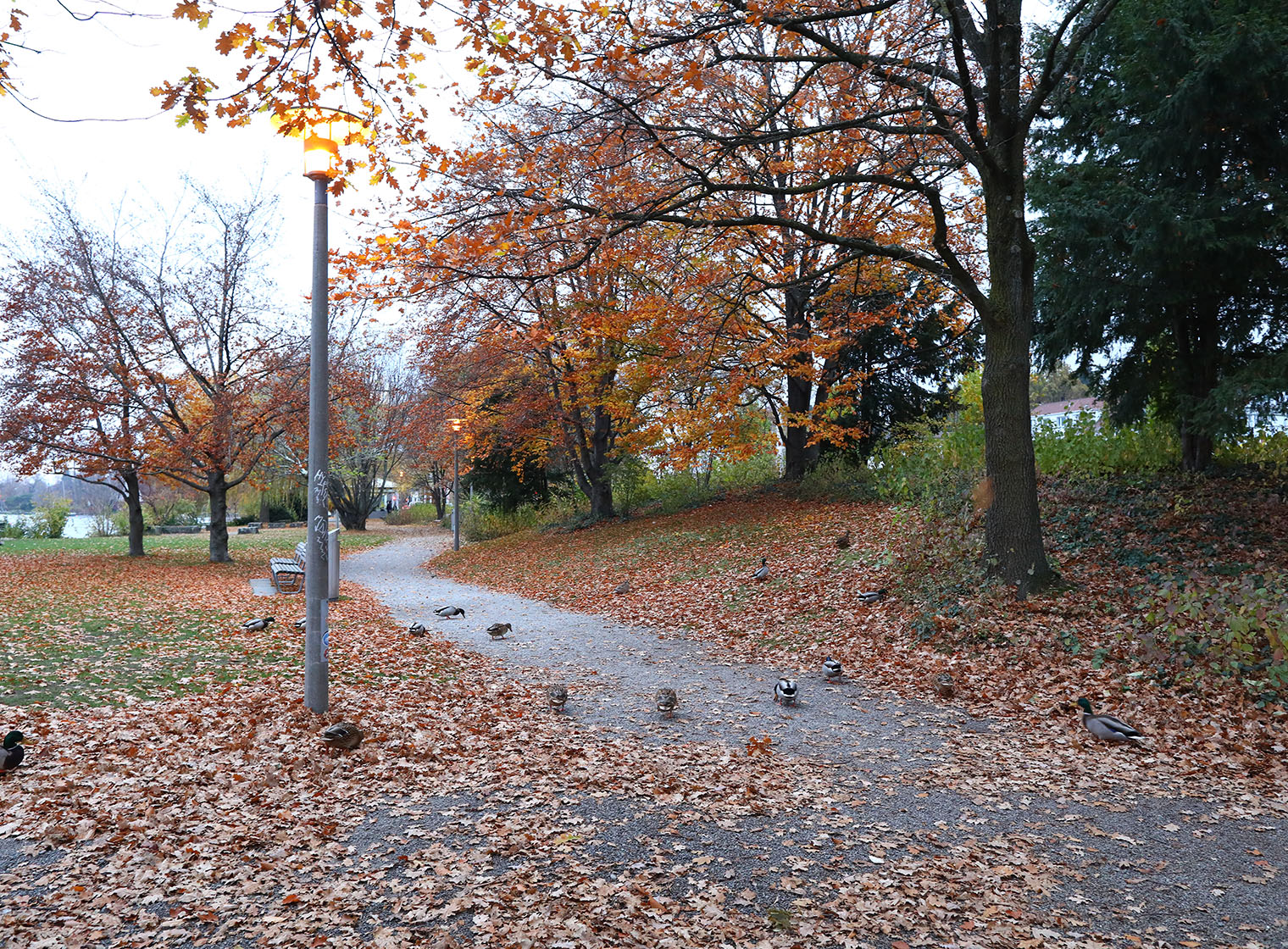} \\
    \includegraphics[width=0.24\linewidth]{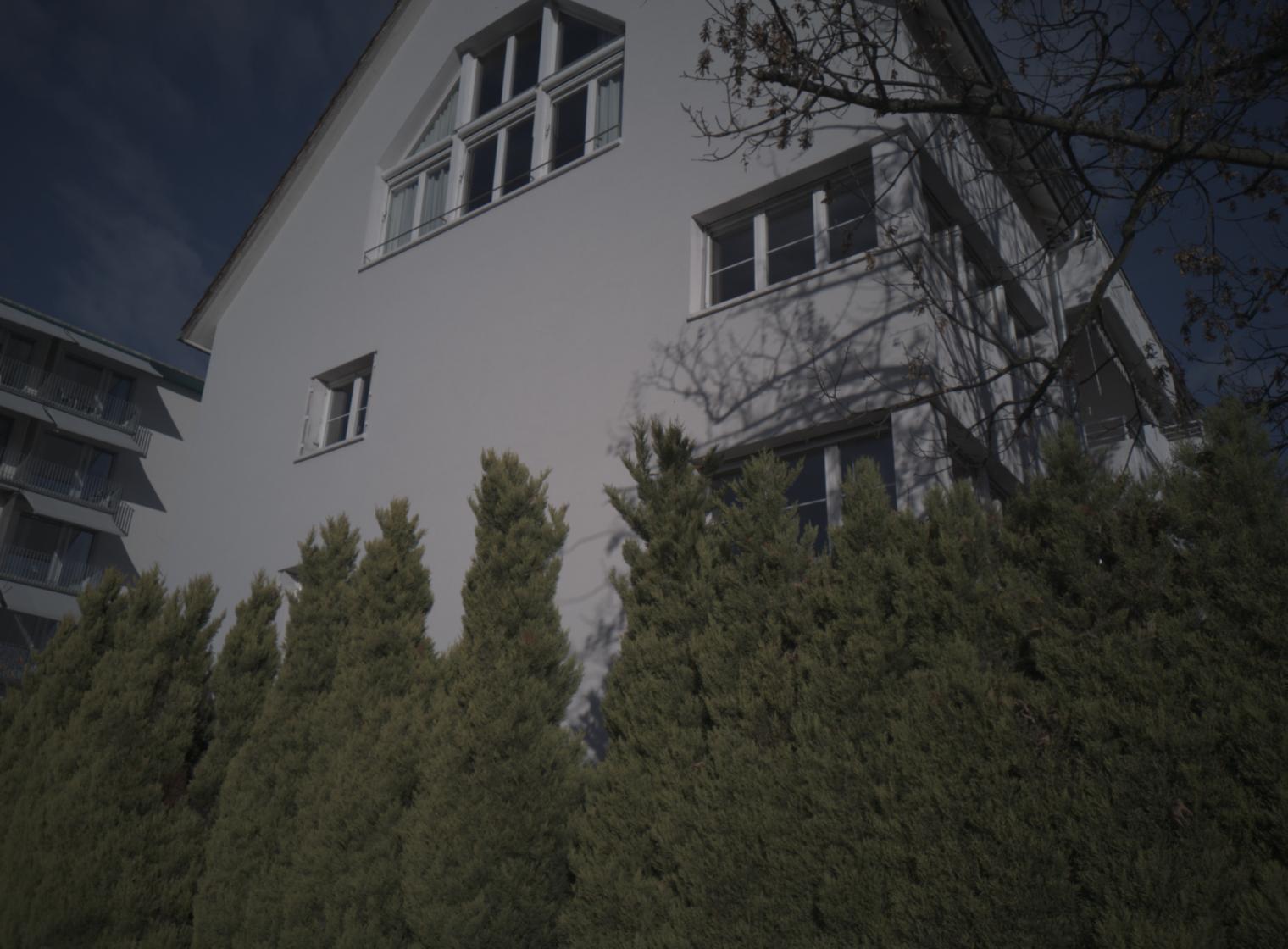}&
    \includegraphics[width=0.24\linewidth]{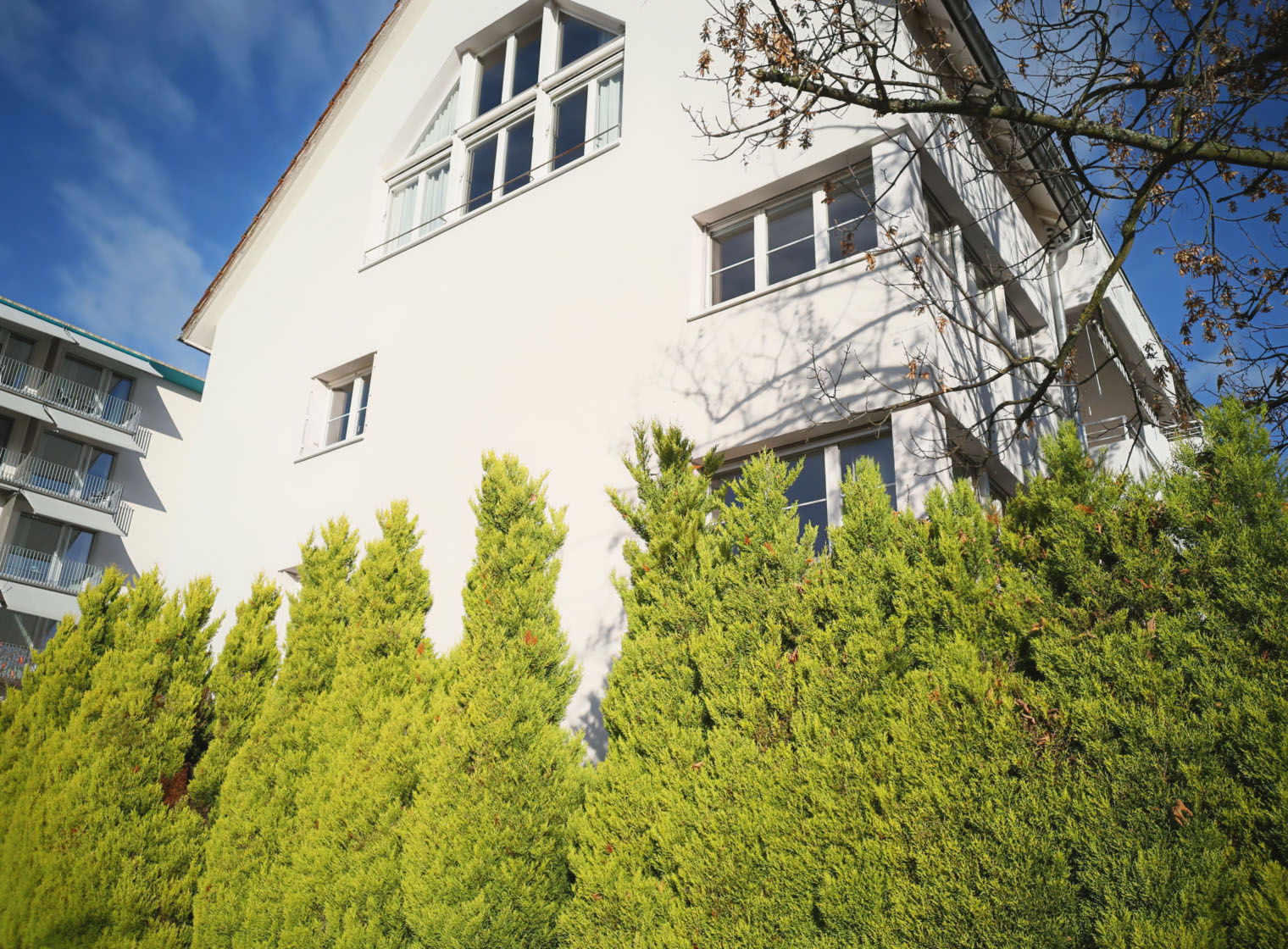}&
    \includegraphics[width=0.24\linewidth]{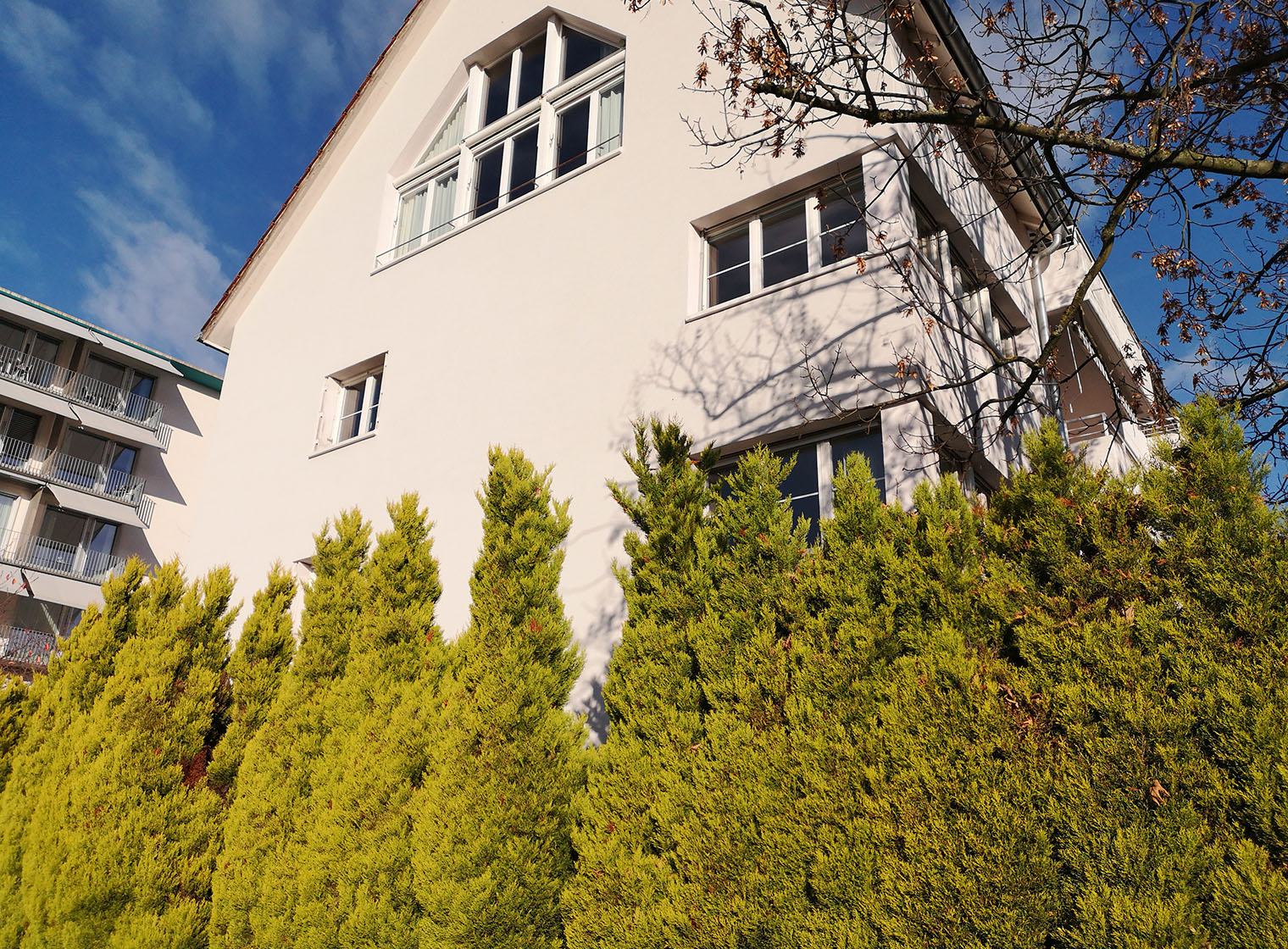}&
    \includegraphics[width=0.24\linewidth]{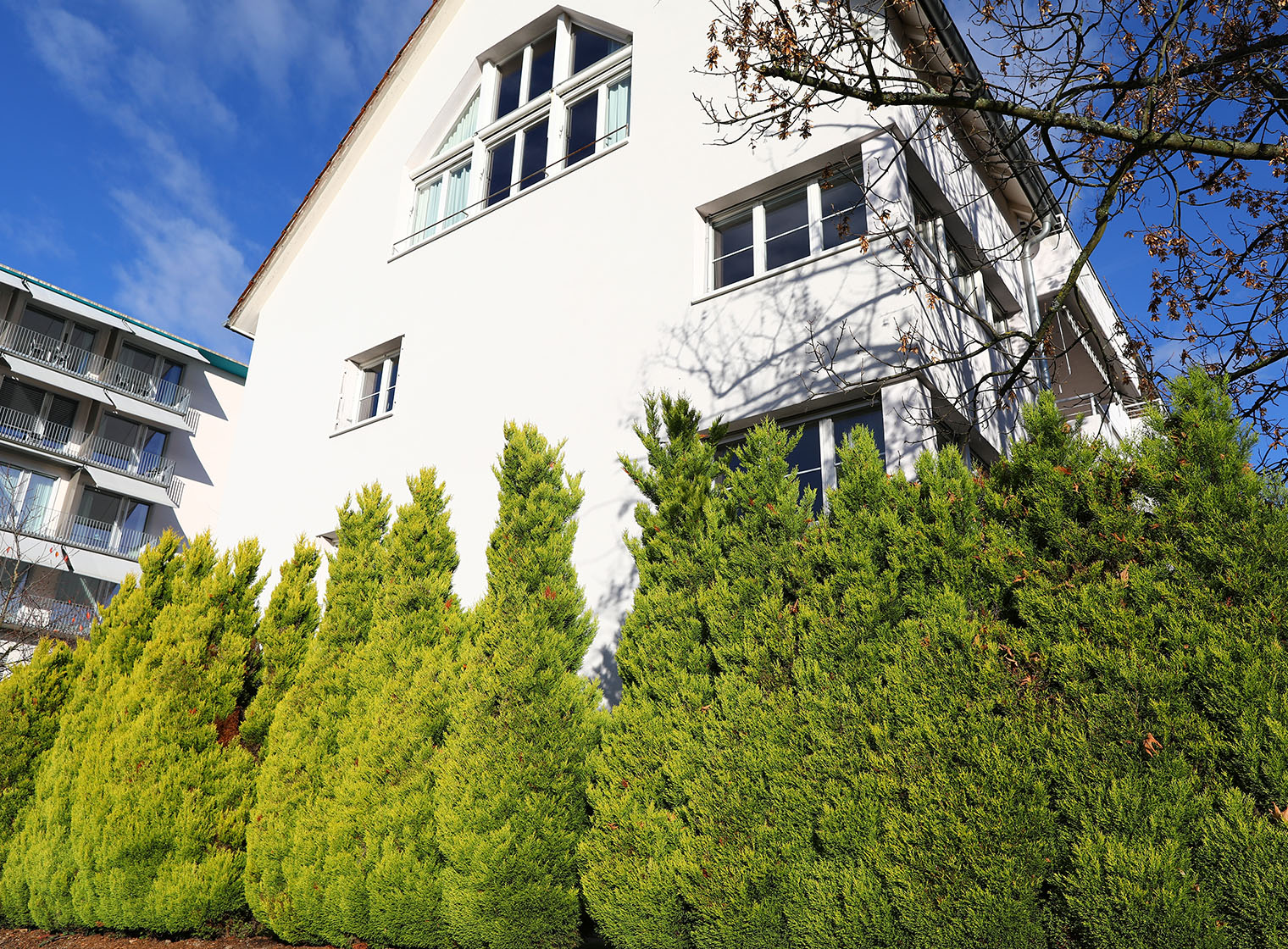} \\
    \includegraphics[width=0.24\linewidth]{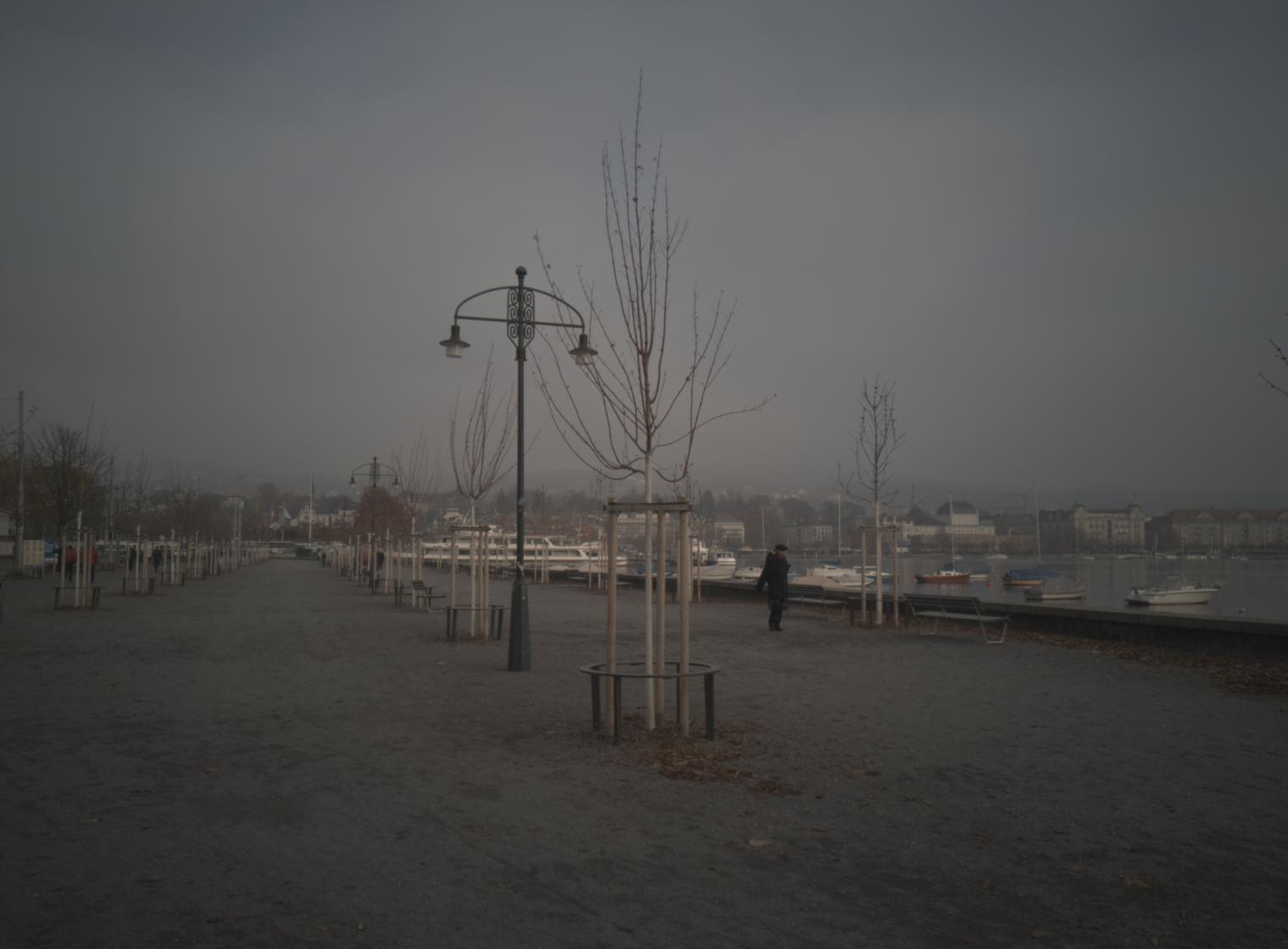}&
    \includegraphics[width=0.24\linewidth]{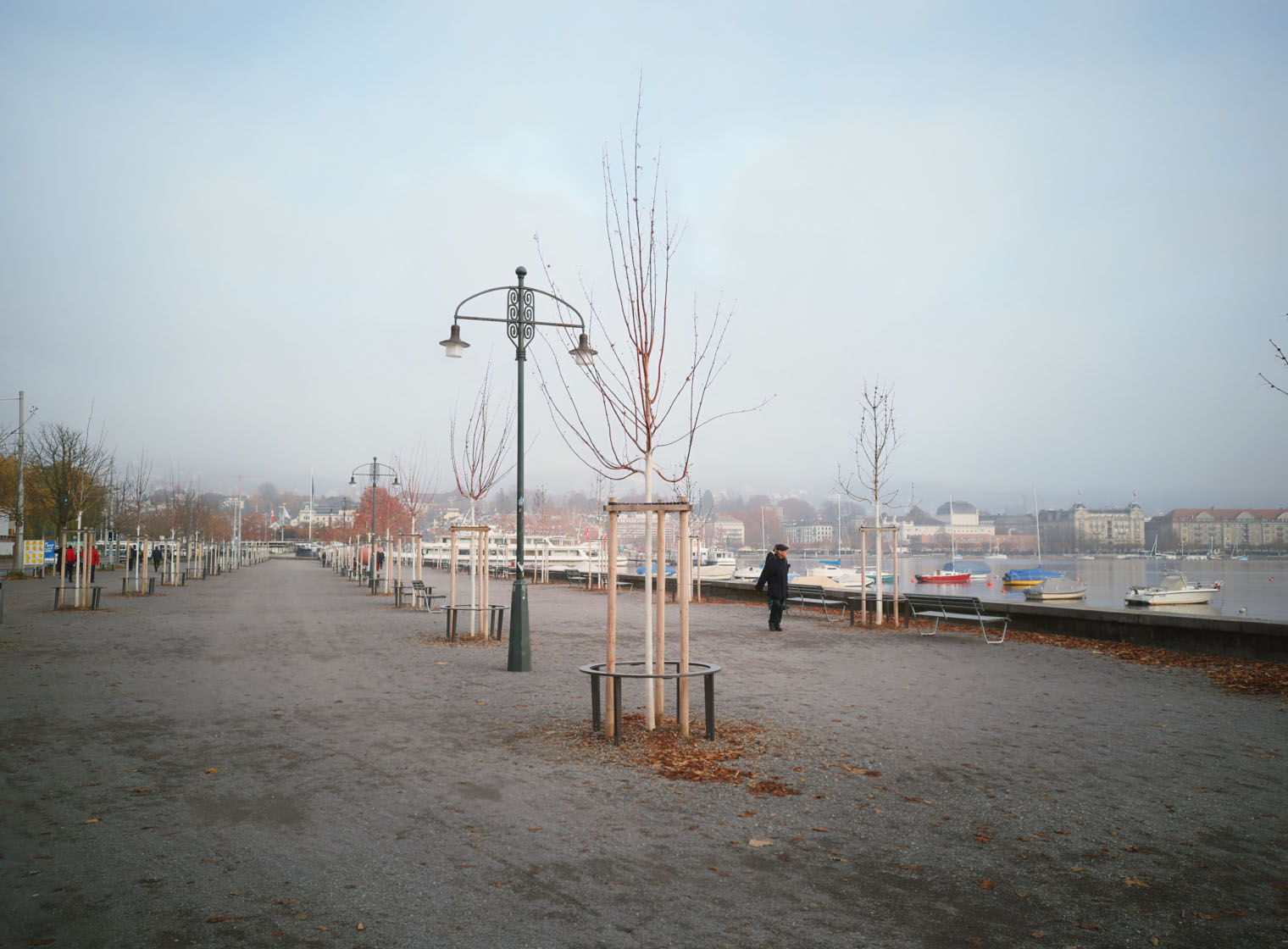}&
    \includegraphics[width=0.24\linewidth]{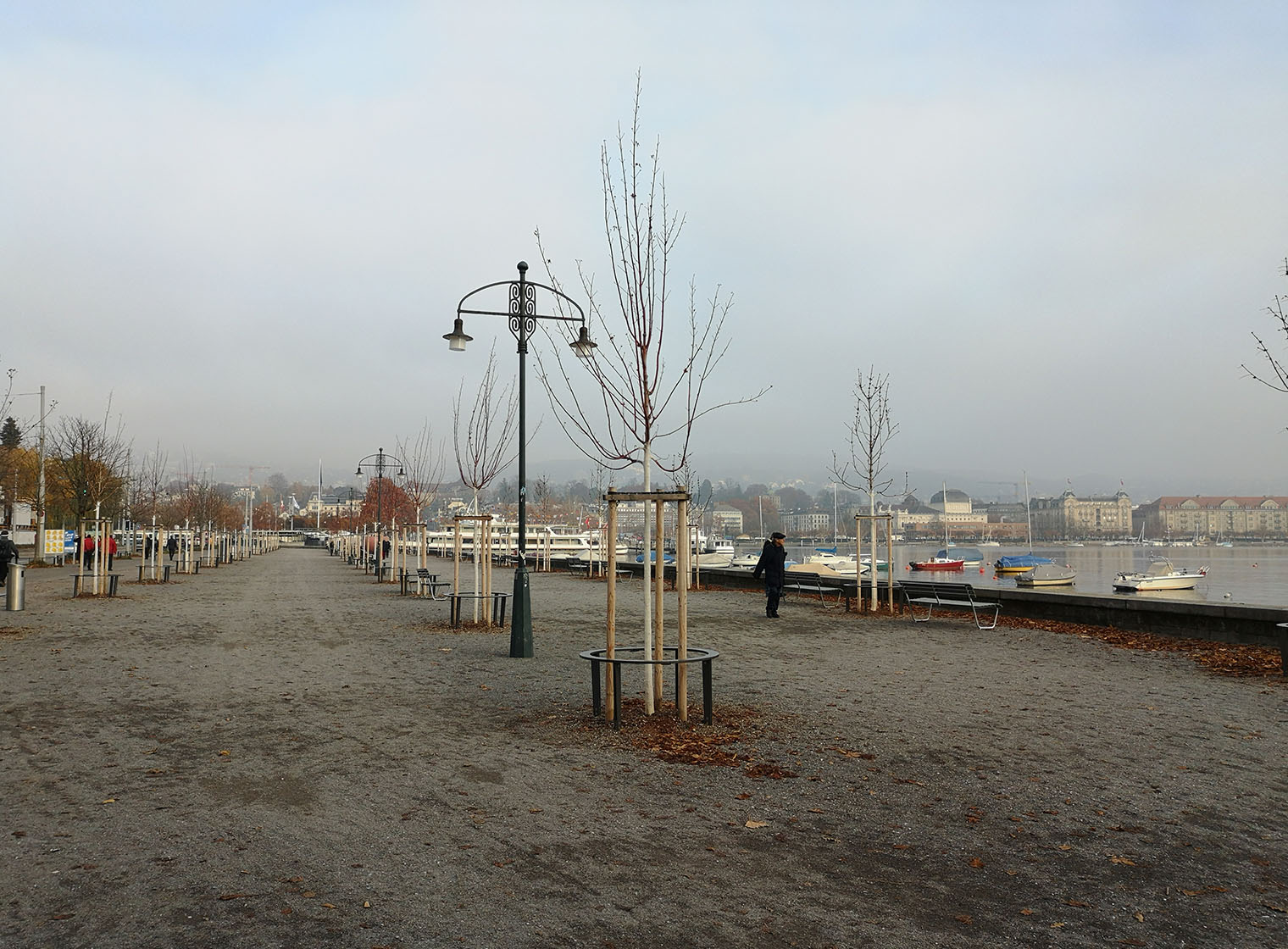}&
    \includegraphics[width=0.24\linewidth]{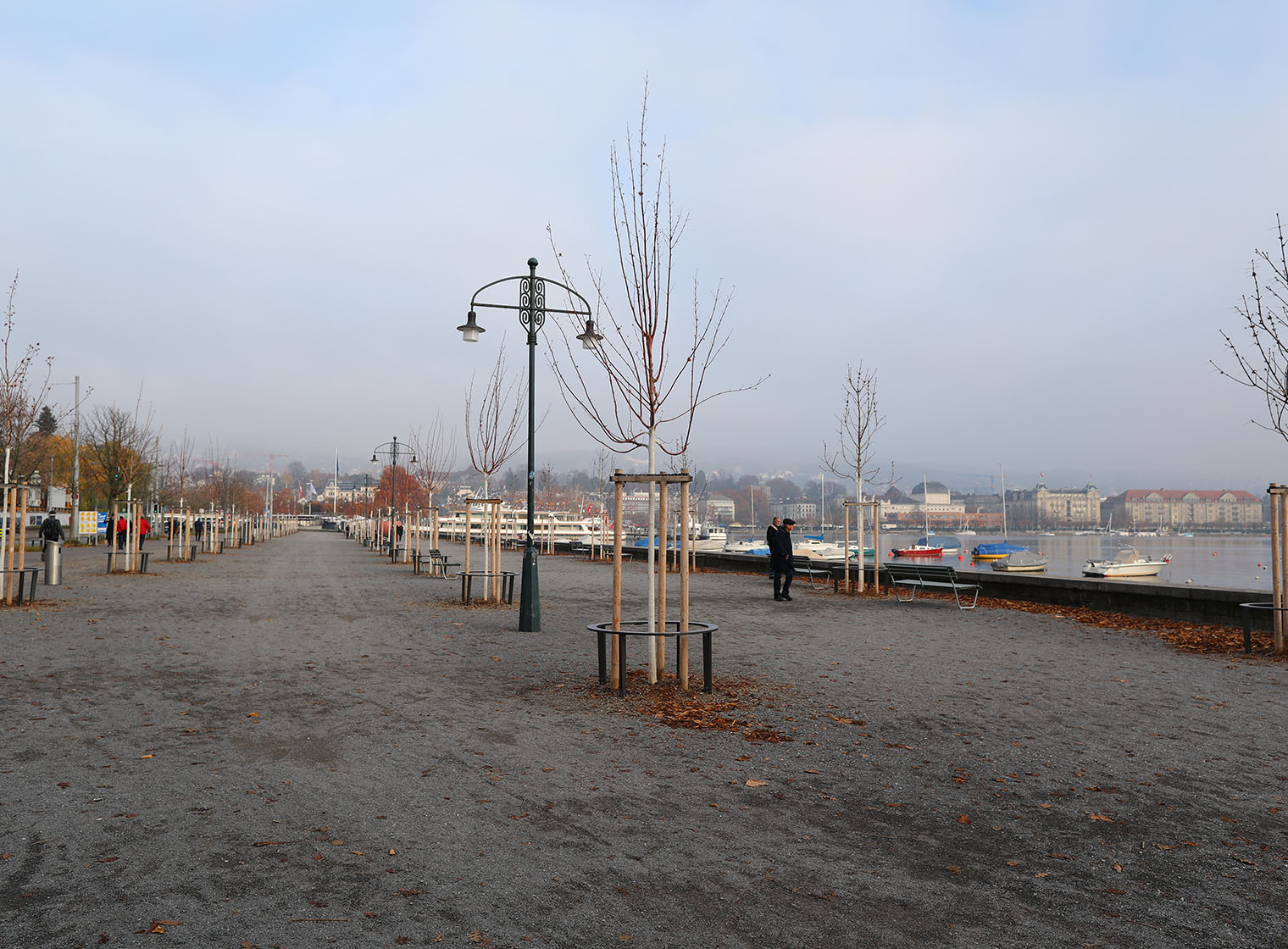} \\
    \includegraphics[width=0.24\linewidth]{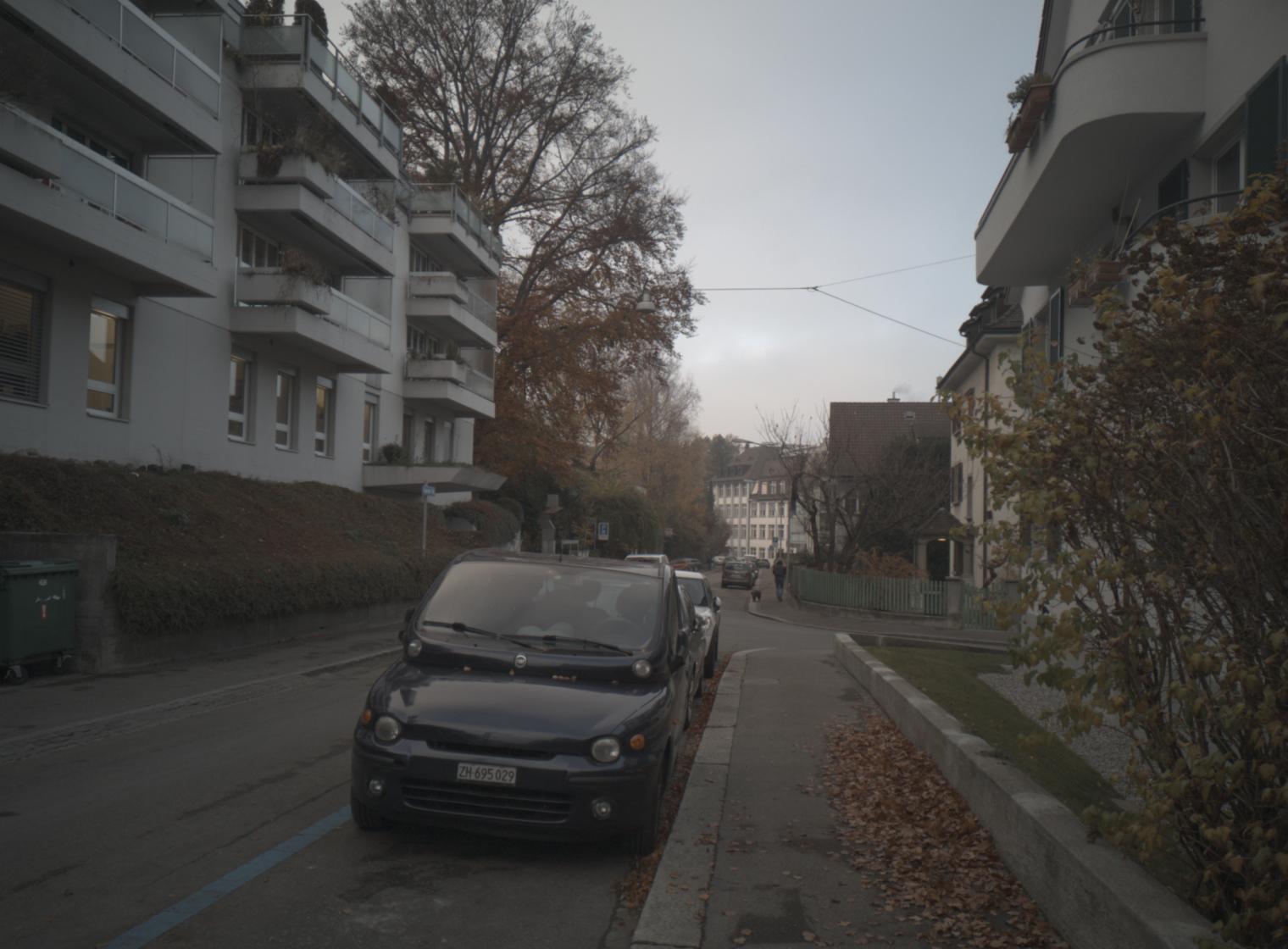}&
    \includegraphics[width=0.24\linewidth]{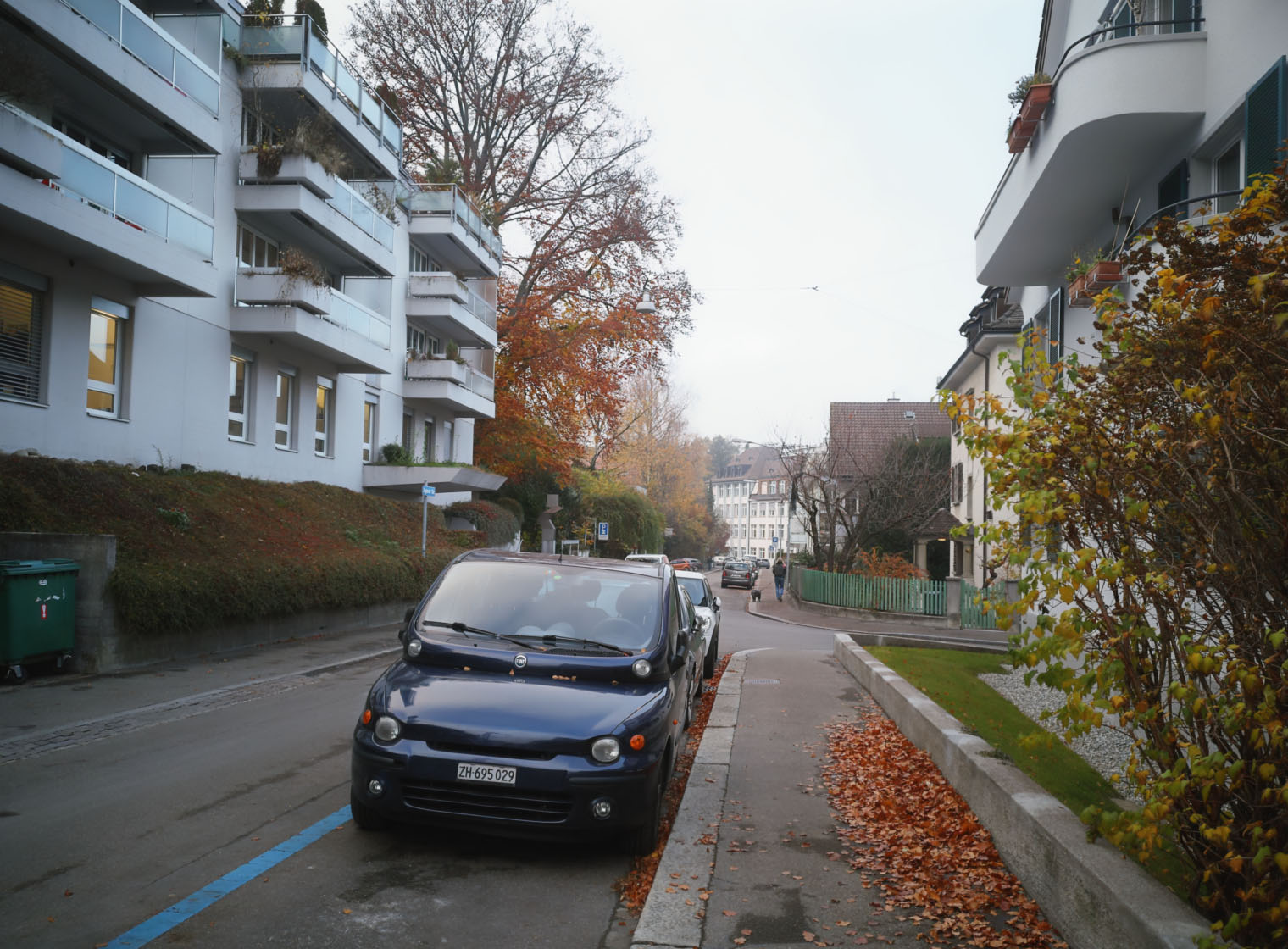}&
    \includegraphics[width=0.24\linewidth]{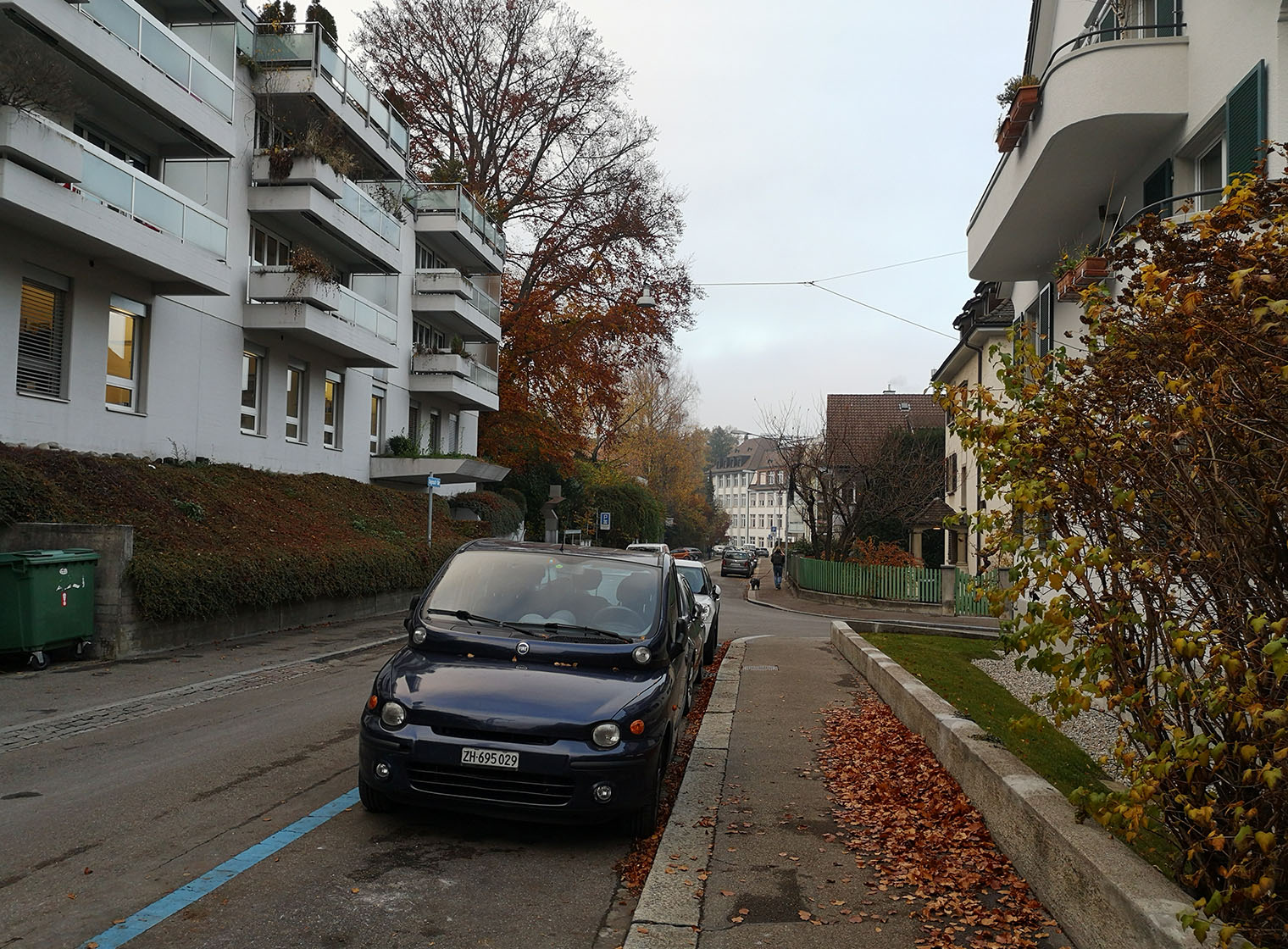}&
    \includegraphics[width=0.24\linewidth]{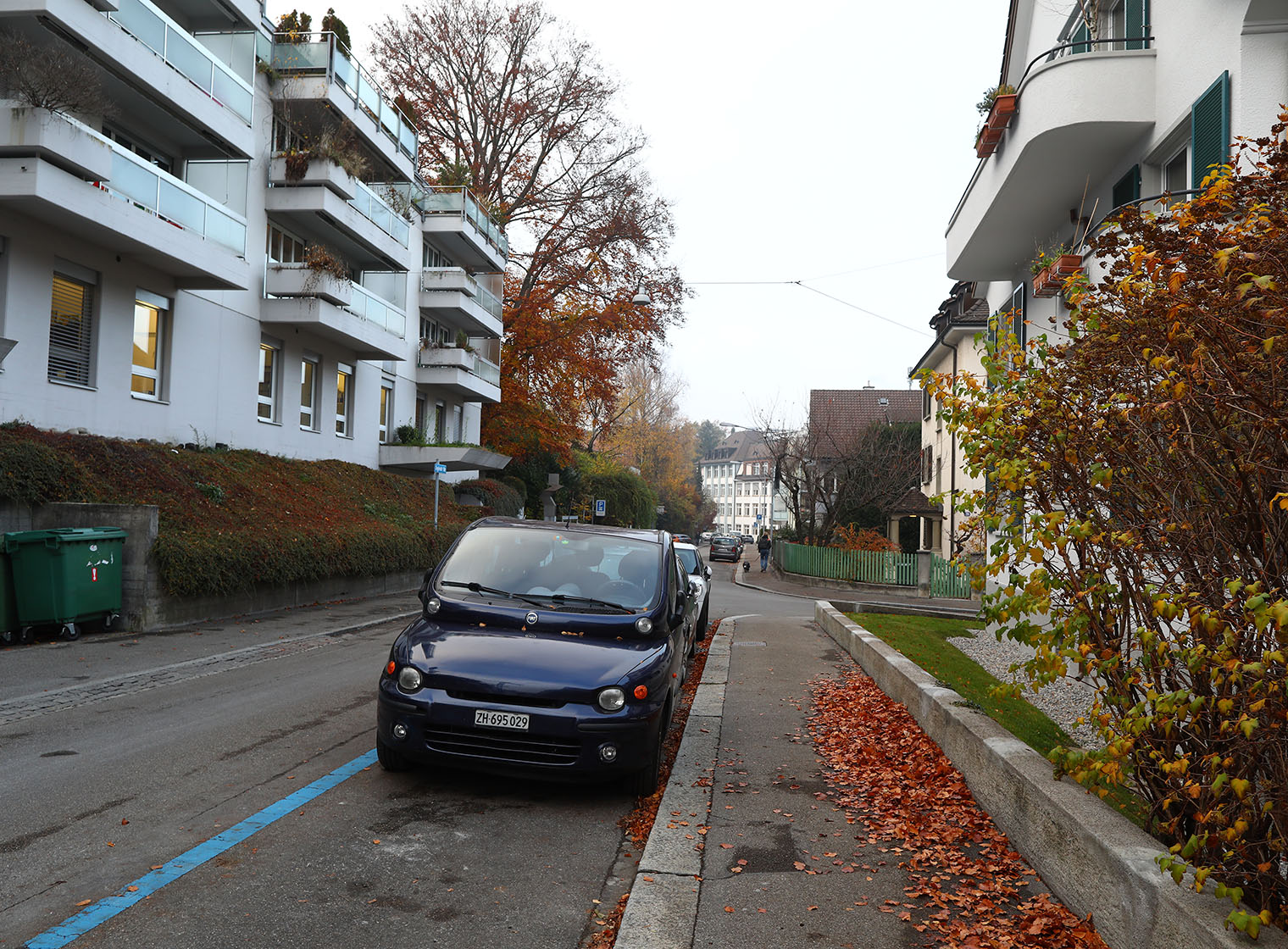} \\
\end{tabular}
}
\vspace{0.0cm}
\caption{\small{Sample visual results obtained with the proposed deep learning method. Best zoomed on screen.}}
\label{fig:sample_results}
\vspace{-0.2cm}
\end{figure*}

\subsection{Loss functions}

The loss function used to train the model depends on the corresponding level / scale of the produced images:

\smallskip

\textBF{Levels 4-5} operate with images downscaled by a factor of 8 and 16, respectively, therefore they are mainly targeted at global color and brightness / gamma correction. These layers are trained to minimize the mean squared error (MSE) since the perceptual losses are not efficient at these scales.

\smallskip

\textBF{Levels 2-3} are processing 2x / 4x downscaled images, and are mostly working on the global content domain. The goal of these layers is to refine the color / shape properties of various objects on the image, taking into account their semantic meaning. They are trained with a combination of the VGG-based~\cite{johnson2016perceptual} perceptual and MSE loss functions taken in the ratio of 4:1.

\smallskip

\textBF{Level 1} is working on the original image scale and is primarily trained to perform local image corrections: texture enhancement, noise removal, local color processing, \etc., while using the results obtained from the lower layers. It is trained using the following loss function:
\begin{equation*}
\label{eq:loss}
\mathcal{L}_{\text{Level 1}} = \mathcal{L}_{\text{VGG}} + 0.75 \cdot \mathcal{L}_{\text{SSIM}} + 0.05 \cdot \mathcal{L}_{\text{MSE}},
\end{equation*}
where the value of each loss is normalized to 1. The structural similarity (SSIM) loss~\cite{wang2003multiscale} is used here to increase the dynamic range of the reconstructed photos, while the MSE loss is added to prevent significant color deviations.

\smallskip

The above coefficients were chosen based on the results of the preliminary experiments on the considered RAW to RGB dataset. We should emphasize that each level is trained together with all (already pre-trained) lower levels to ensure a deeper connection between the layers.

\vspace{-1.8mm}
\subsection{Technical details}

The model was implemented in TensorFlow~\footnote{\url{https://github.com/aiff22/pynet}} and was trained on a single \textit{Nvidia Tesla V100} GPU with a batch size ranging from 10 to 50 depending on the training scale. The parameters of the model were optimized for 5 $\sim$ 20 epochs using Adam~\cite{kingma2014adam} algorithm with a learning rate of $5e-5$. The entire PyNET model consists of 47.5M parameters, and it takes 3.8 seconds to process one 12MP photo (2944$\times$3958 pixels) on the above mentioned GPU.

\begin{figure*}[t!]
\centering
\setlength{\tabcolsep}{1pt}
\resizebox{\linewidth}{!}
{
\begin{tabular}{ccccc}
    \includegraphics[width=0.2\linewidth]{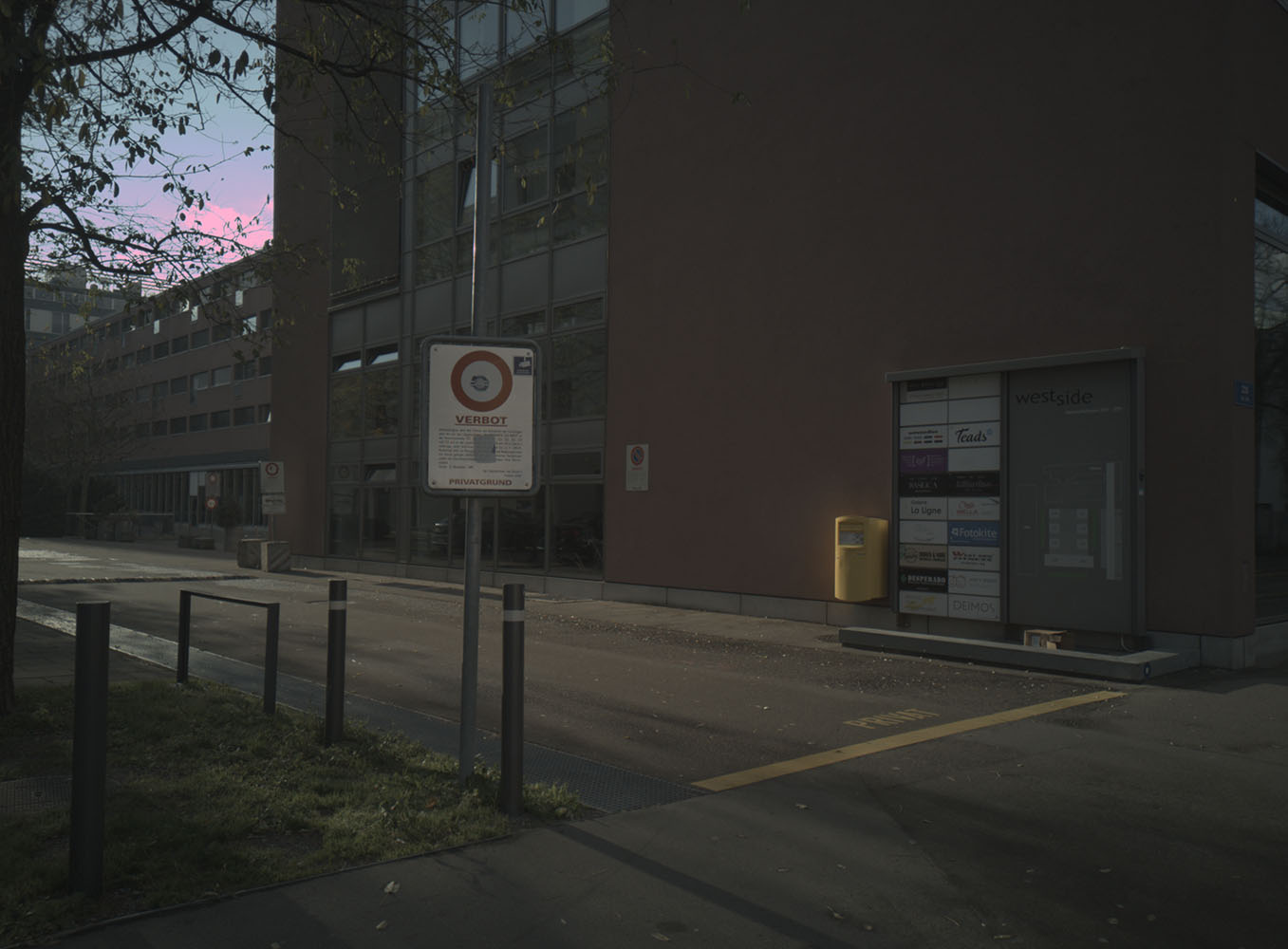}&
    \includegraphics[width=0.2\linewidth]{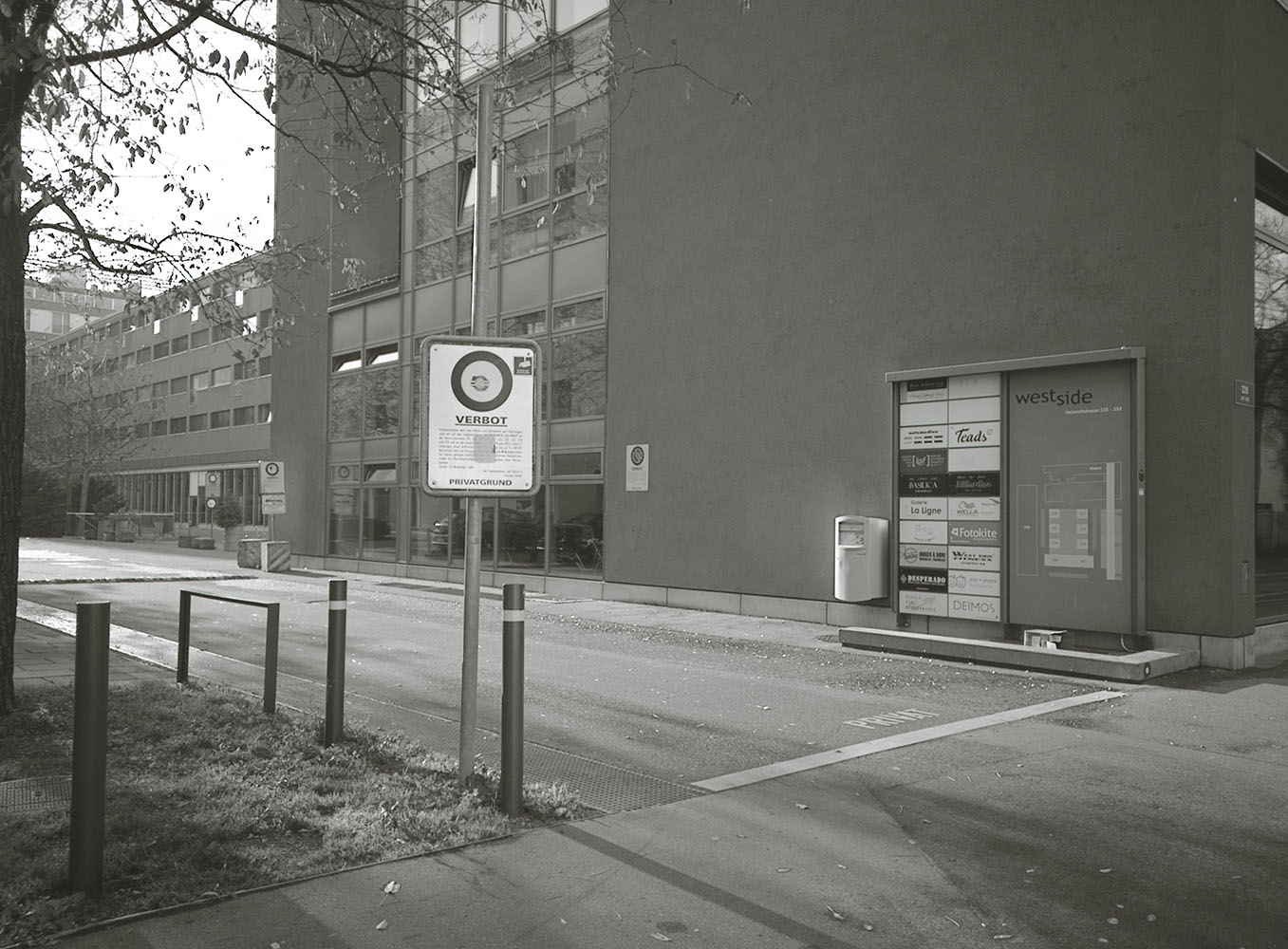}&
    \includegraphics[width=0.2\linewidth]{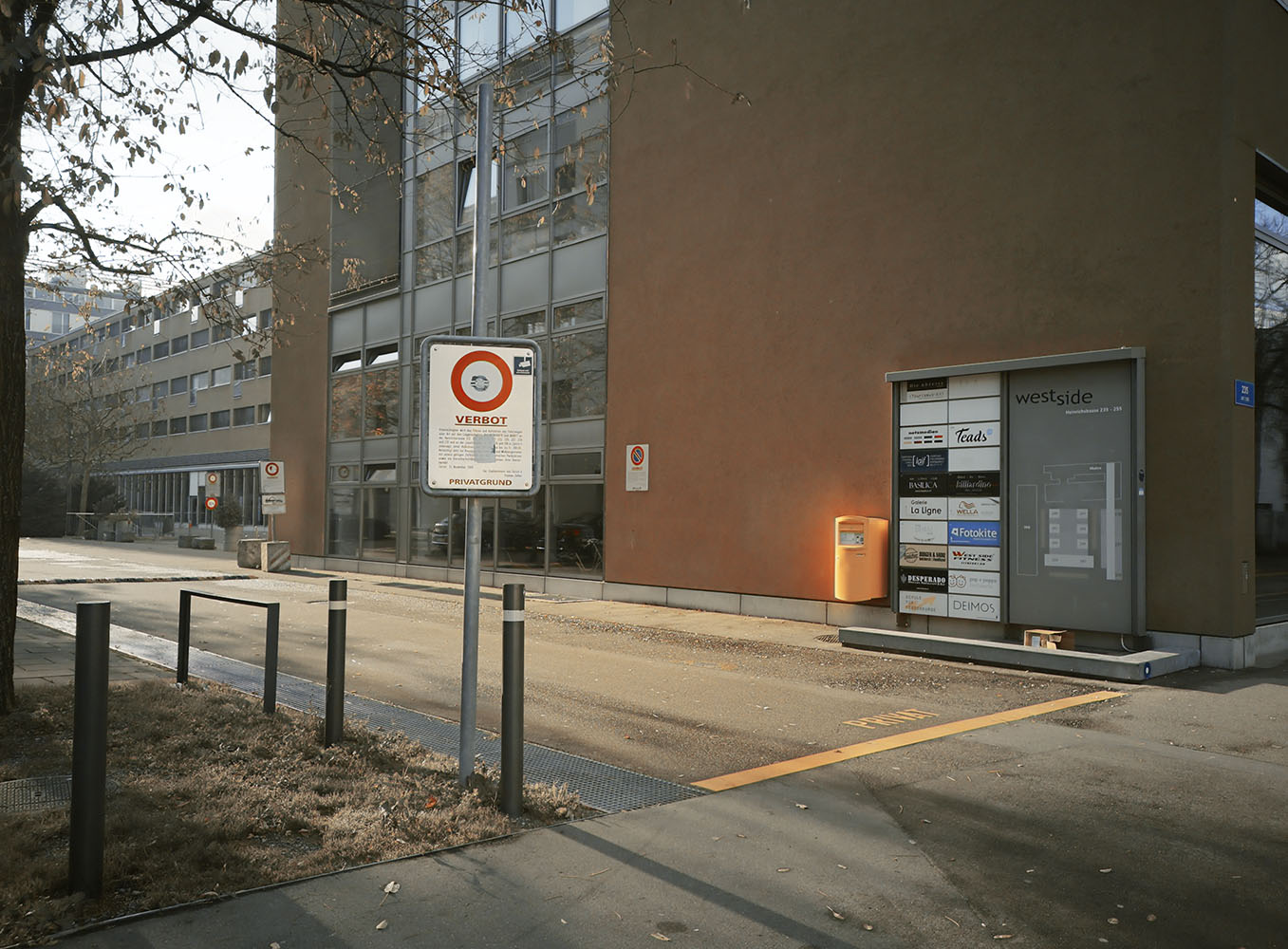}&
    \includegraphics[width=0.2\linewidth]{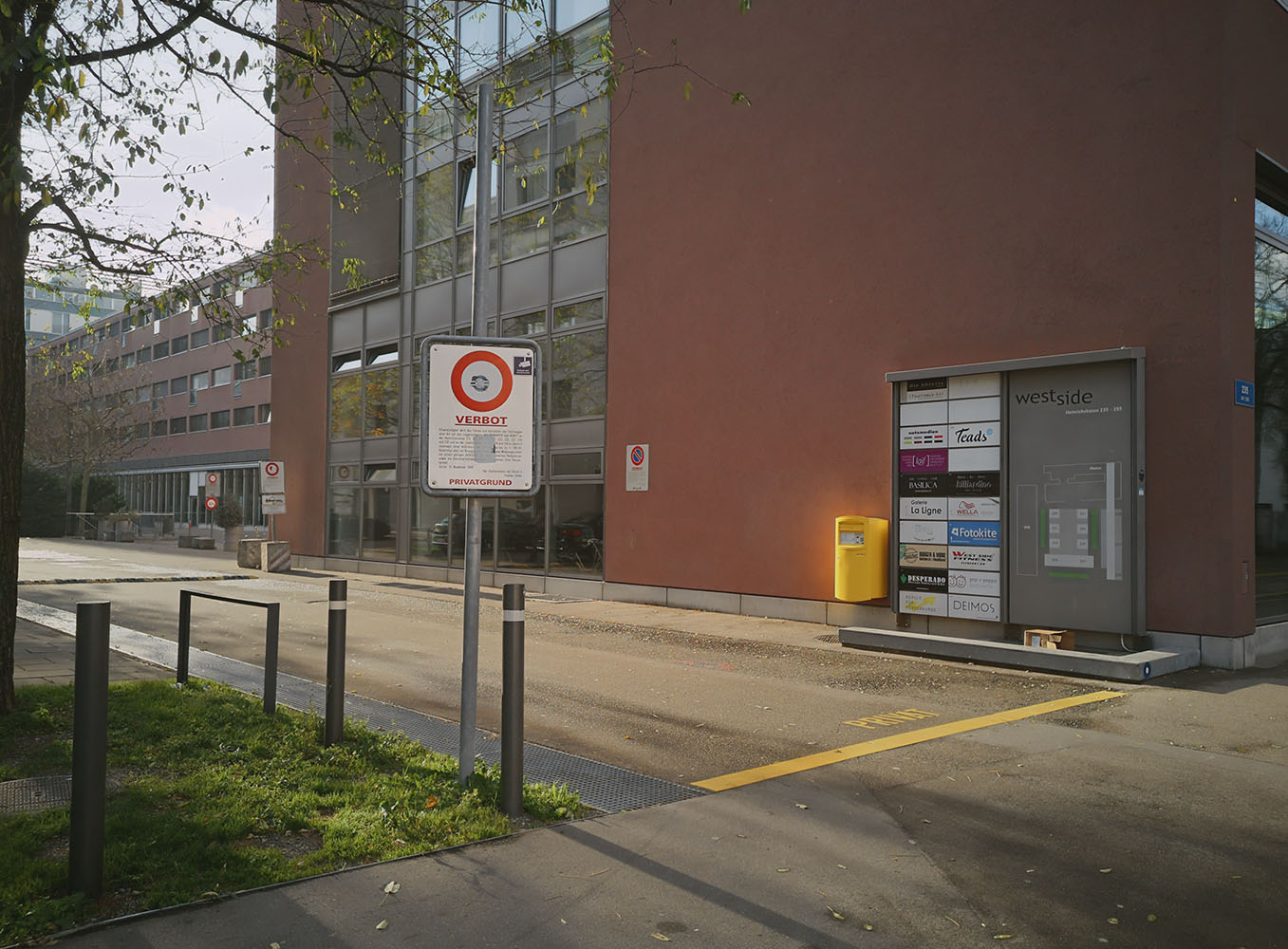}&
    \includegraphics[width=0.2\linewidth]{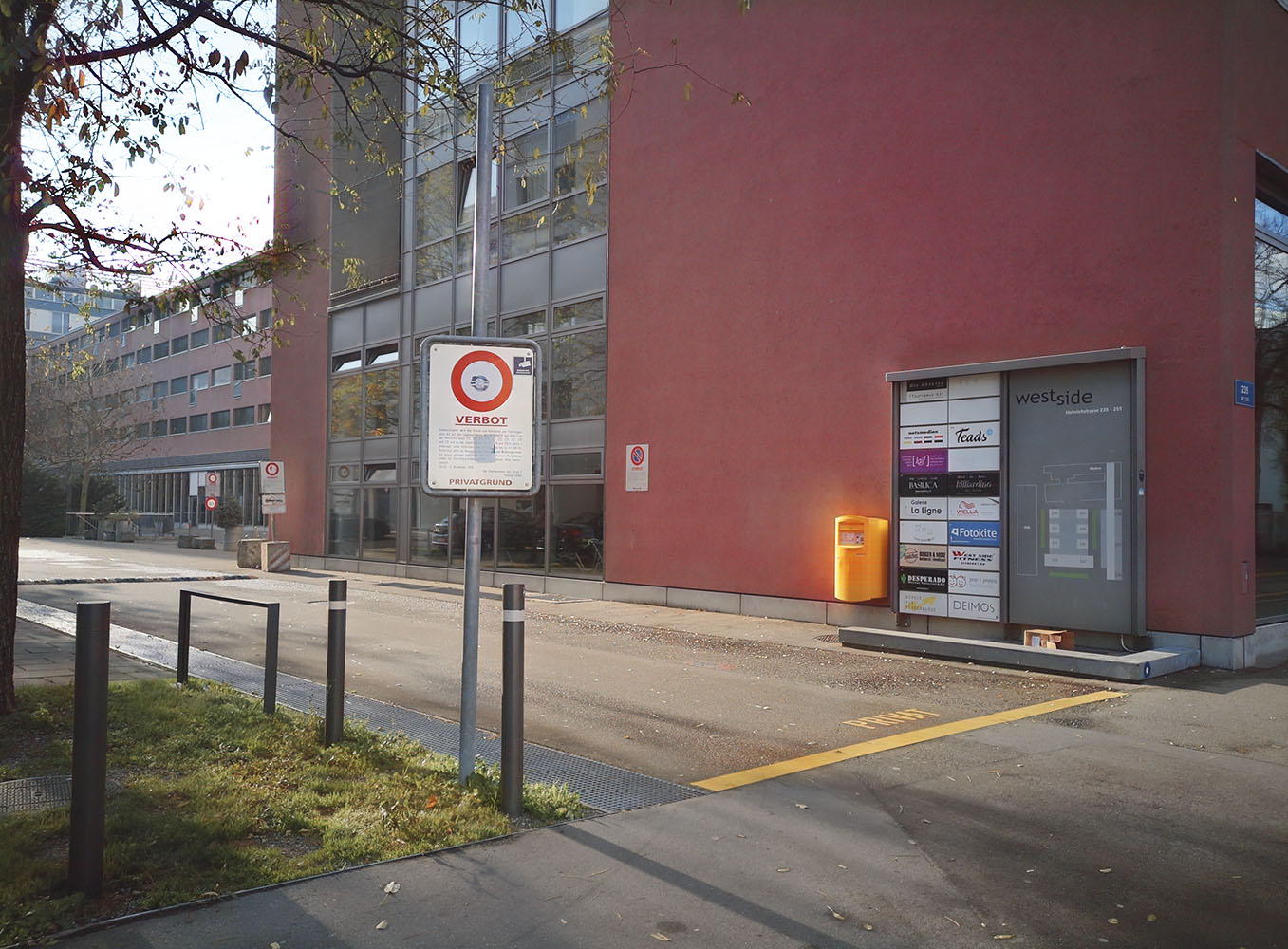} \\
    \includegraphics[width=0.2\linewidth]{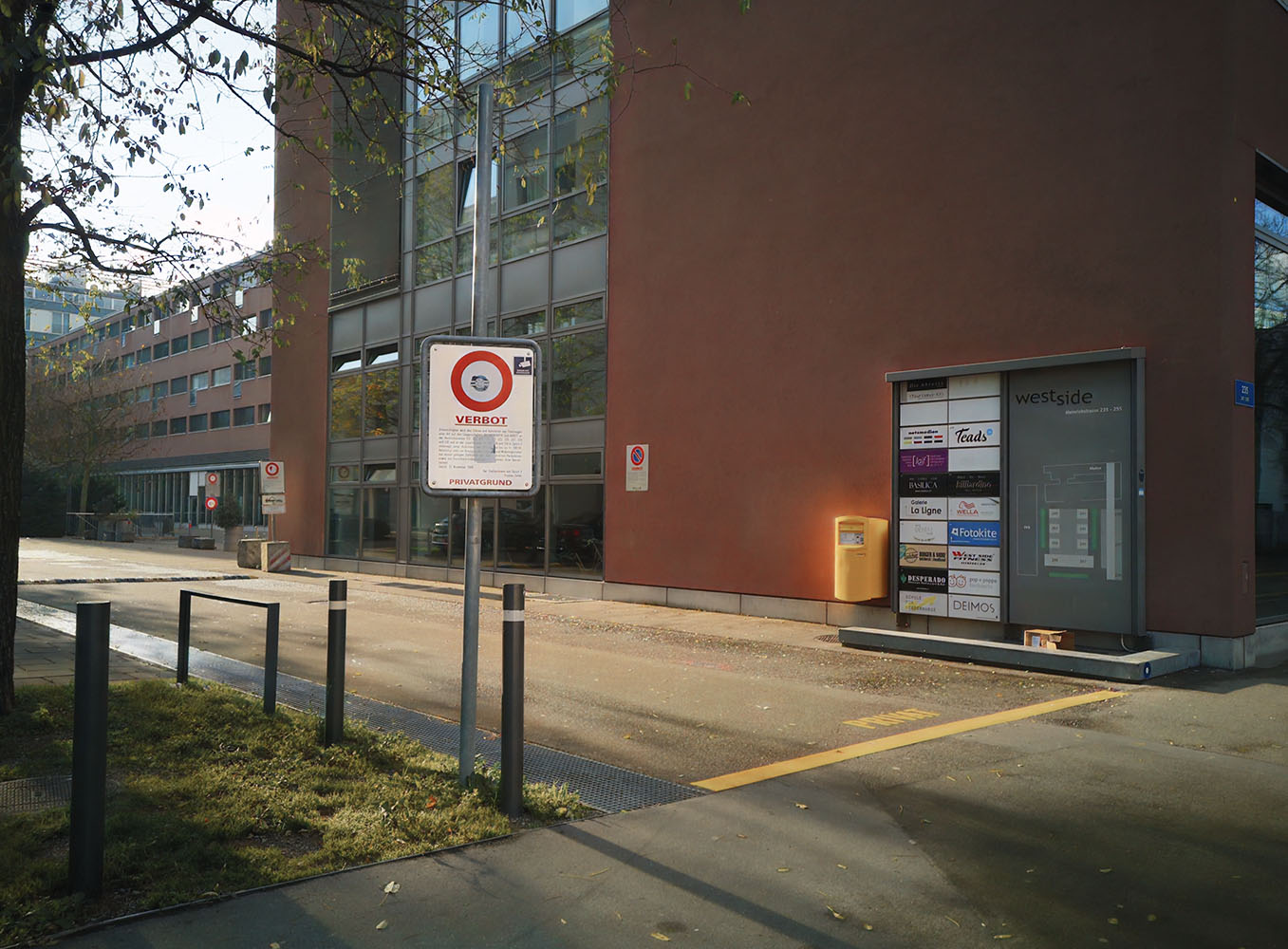}&
    \includegraphics[width=0.2\linewidth]{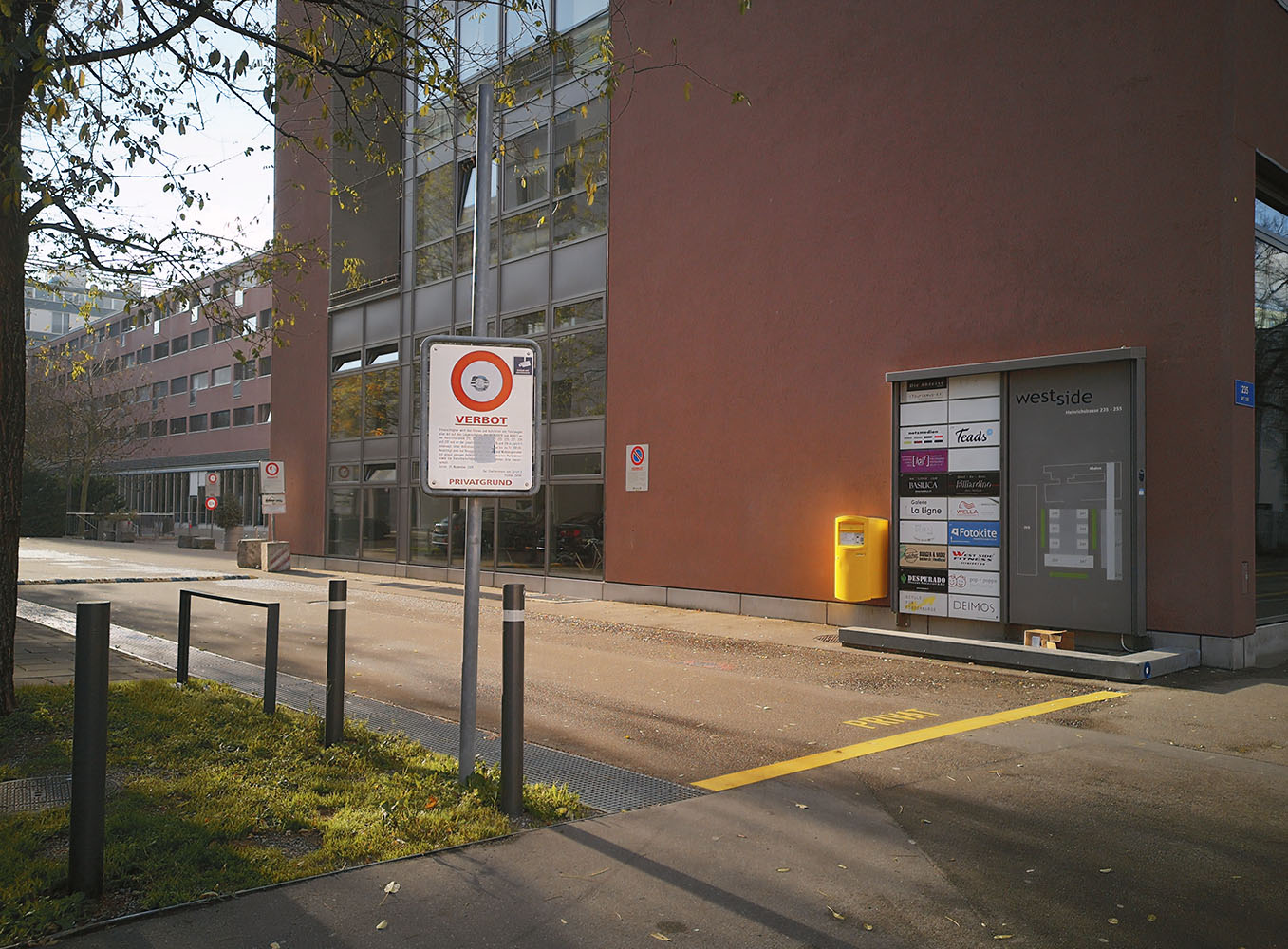}&
    \includegraphics[width=0.2\linewidth]{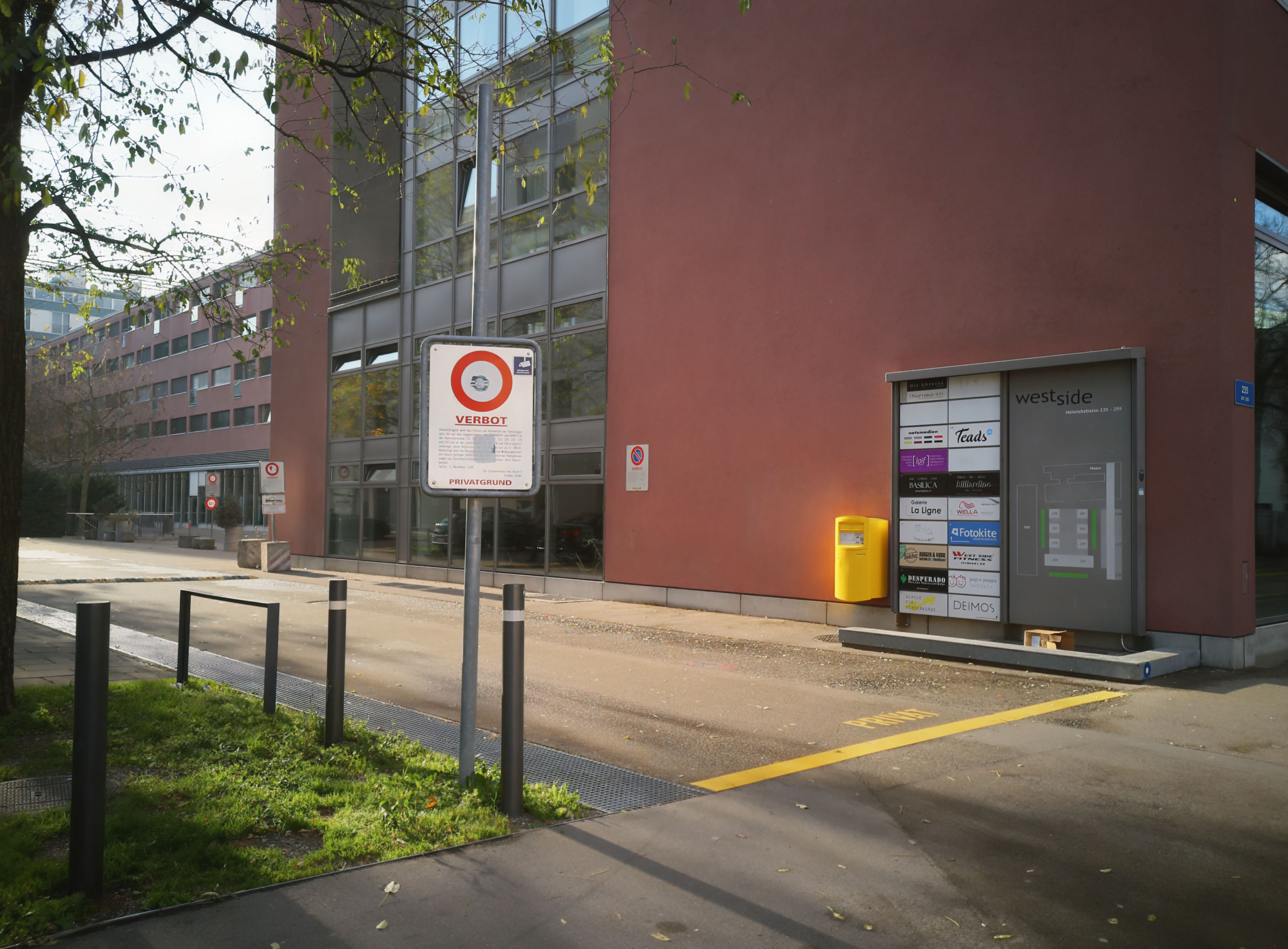}&
    \includegraphics[width=0.2\linewidth]{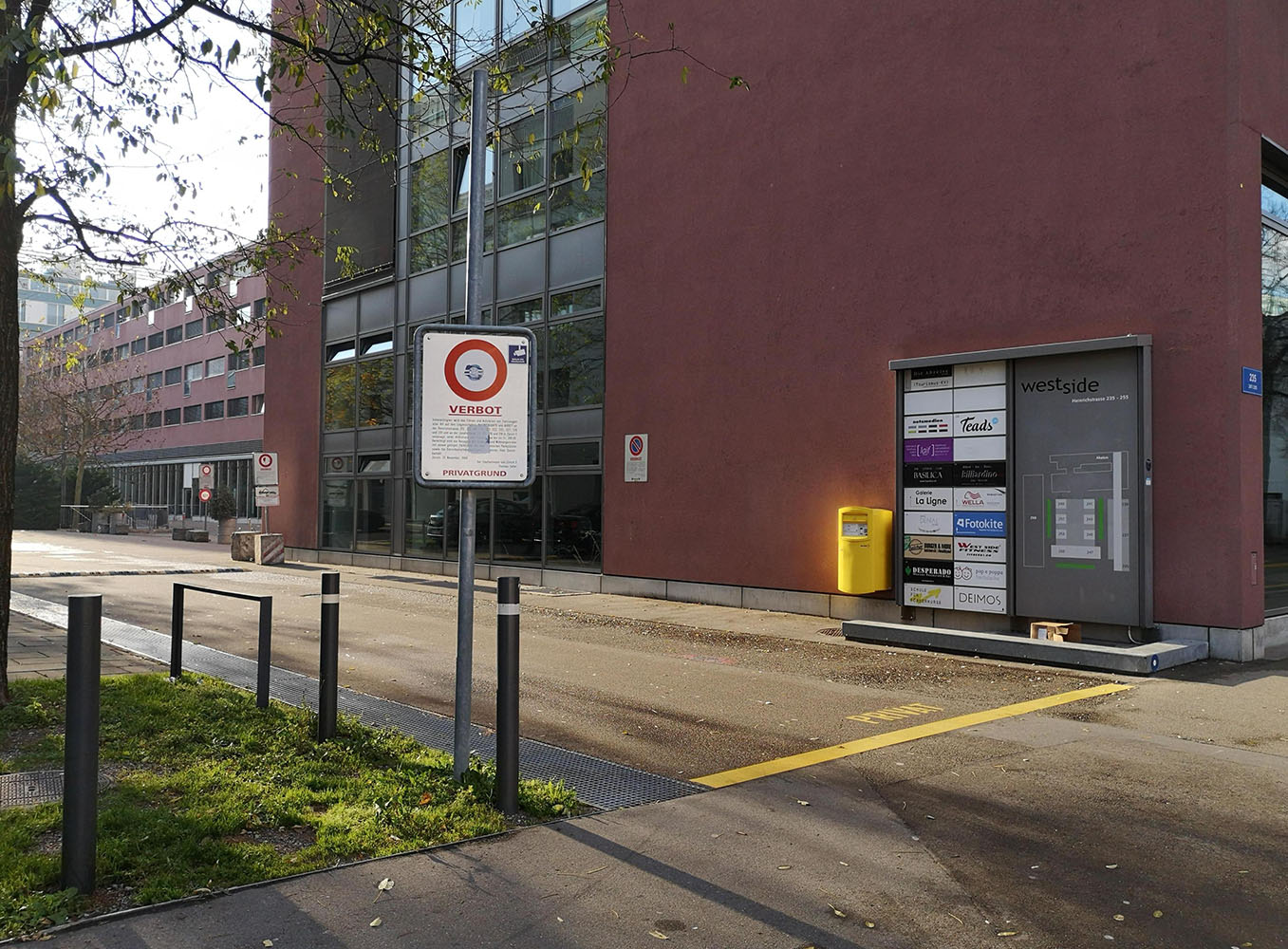}&
    \includegraphics[width=0.2\linewidth]{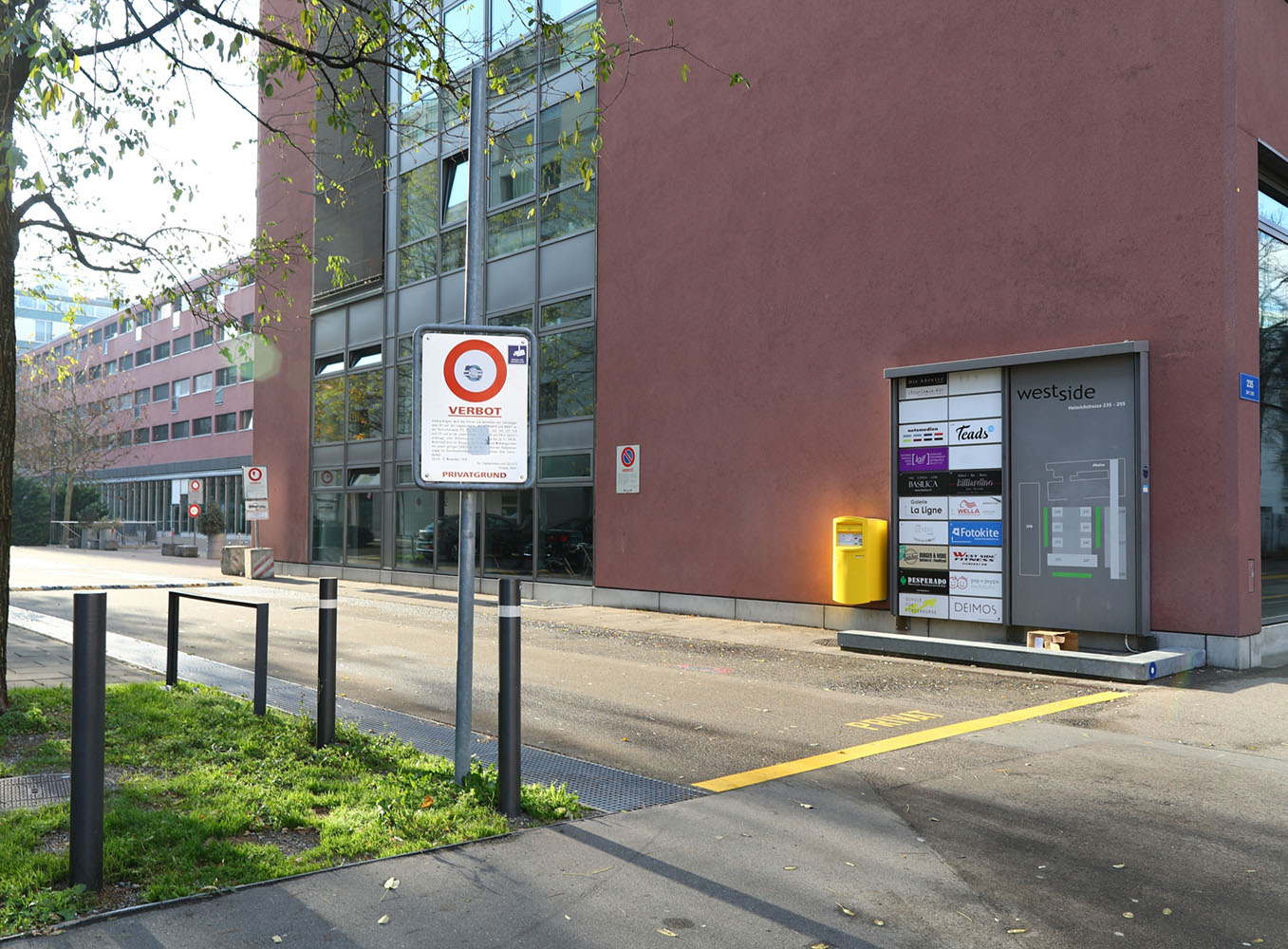} \\
\end{tabular}
}
\vspace{0.0cm}
\caption{\small{Visual results obtained with 7 different architectures. From left to right, top to bottom: visualized RAW photo, SRCNN~\cite{dong2016image}, VDSR~\cite{kim2016accurate}, SRGAN~\cite{ledig2017photo}, Pix2Pix~\cite{isola2017image}, U-Net~\cite{ronneberger2015u}, DPED~\cite{ignatov2017dslr}, our PyNET architecture, Huawei ISP image and the target Canon photo.}}
\label{fig:photo_comparison}
\vspace{-0.2cm}
\end{figure*}

\section{Experiments}

In this section, we evaluate the quantitative and qualitative performance of the proposed solution on the real RAW to RGB mapping problem. In particular, our goal is to answer the following three questions:

\begin{itemize}
\item How well the proposed approach performs numerically and perceptually compared to common deep learning models widely used for various image-to-image mapping problems.
\item How good is the quality of the reconstructed images in comparison to the built-in ISP system of the Huawei P20 camera phone.
\item Is the proposed solution generalizable to other mobile phones / camera sensors.
\end{itemize}

To answer these questions, we trained a wide range of deep learning models including the SPADE~\cite{park2019semantic}, DPED~\cite{ignatov2017dslr}, U-Net~\cite{ronneberger2015u}, Pix2Pix~\cite{isola2017image}, SRGAN~\cite{ledig2017photo}, VDSR~\cite{kim2016accurate} and SRCNN~\cite{dong2016image} on the same data and measured the obtained results. We performed a user study involving a large number of participants asked to rate the target DSLR photos, the photos obtained with P20's ISP pipeline and the images reconstructed with our method. Finally, we applied our pre-trained model to RAW photos from a different device~-- BlackBerry KeyOne smartphone, to see if the considered approach is able to reconstruct RGB images when using camera sensor data obtained with other hardware. The results of these experiments are described in detail in the following three sections.

\subsection{Quantitative evaluation}

Before starting the comparison, we first trained the proposed PyNET model and performed a quick inspection of the produced visual results. An example of the reconstructed images obtained with the proposed model is shown in Figure~\ref{fig:sample_results}. The produced RGB photos do not contain any notable artifacts or corruptions at both the local and global levels, and the only major issue is vignetting caused by camera optics. Compared to photos obtained with Huawei's ISP, the reconstructed images have brighter colors and more natural local texture, while their sharpness is slightly lower, which is visible when looking at zoomed-in images. We expect that this might be caused by P20's second 20 MP monochrome camera sensor that can be used for image sharpening. In general, the overall quality of photos obtained with Huawei's ISP and reconstructed with our method is quite comparable, though both of them are worse than the images produced by the Canon 5D DSLR in terms of the color and texture quality.

Next, we performed a quantitative evaluation of the proposed method and alternative deep learning approaches. Table~\ref{tab:scores} shows the resulting PSNR and MS-SSIM scores obtained with different deep learning architecture on the test subset of the considered RAW to RGB mapping dataset. All models were trained twice: with the original loss function and the one used for PyNET training, and the best result was selected in each case. As one can see, PyNET CNN was able to significantly outperform the other models in both the PSNR and MS-SSIM scores. The visual results obtained with these models (Figure~\ref{fig:photo_comparison}) also confirm this conclusion. VGG-19 and SRCNN networks did not have enough power to perform good color reconstruction. The images produced by the SRGAN and U-Net architectures were too dark, with dull colors, while the Pix2Pix had significant problems with accurate color rendering~-- the results are looking unnaturally due to distorted tones. Considerably better image reconstruction was obtained with the DPED model, though in this case the images have a strong yellowish shade and are lacking the vividness. Unfortunately, the SPADE architecture cannot process images of arbitrary resolutions (the size of the input data should be the same as used during the training process), therefore we were unable to generate full images using this method.

\begin{table}
\centering
{
\begin{tabular}{l|c|c}
Method & PSNR & MS-SSIM \\
\hline
PyNET & 21.19 & 0.8620 \\
SPADE~\cite{park2019semantic} & 20.96 & 0.8586 \\
DPED~\cite{ignatov2017dslr} & 20.67 & 0.8560 \\
U-Net~\cite{ronneberger2015u} & 20.81 & 0.8545 \\
Pix2Pix~\cite{isola2017image} & 20.93 & 0.8532 \\
SRGAN~\cite{ledig2017photo} & 20.06 & 0.8501 \\
VDSR~\cite{kim2016accurate} & 19.78 & 0.8457 \\
SRCNN~\cite{dong2016image} & 18.56 & 0.8268 \\
\end{tabular}
}
\vspace{0.0mm}
\caption{\small{Average PSNR/SSIM results on test images.}}
\label{tab:scores}
\vspace{-3.6mm}
\end{table}

\begin{figure*}[t!]
\centering
\setlength{\tabcolsep}{1pt}
\resizebox{\linewidth}{!}
{
\begin{tabular}{cccc}
\tiny{BlackBerry KeyOne RAW Image (Visualized)}\normalsize & \tiny{Reconstructed RGB Image (PyNET)}\normalsize & \tiny{BlackBerry KeyOne ISP Image}\normalsize\\
    \includegraphics[width=0.24\linewidth]{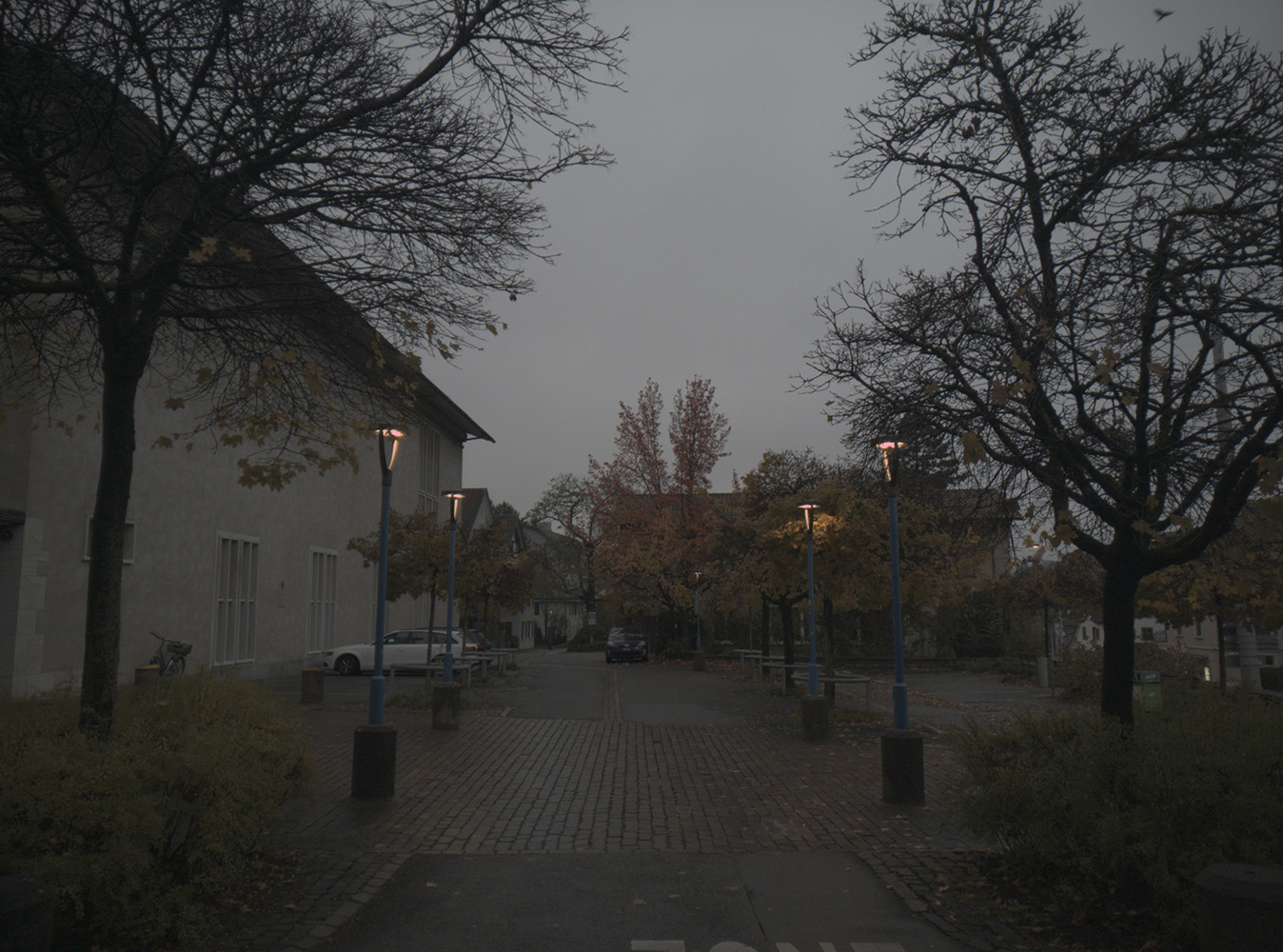}&
    \includegraphics[width=0.24\linewidth]{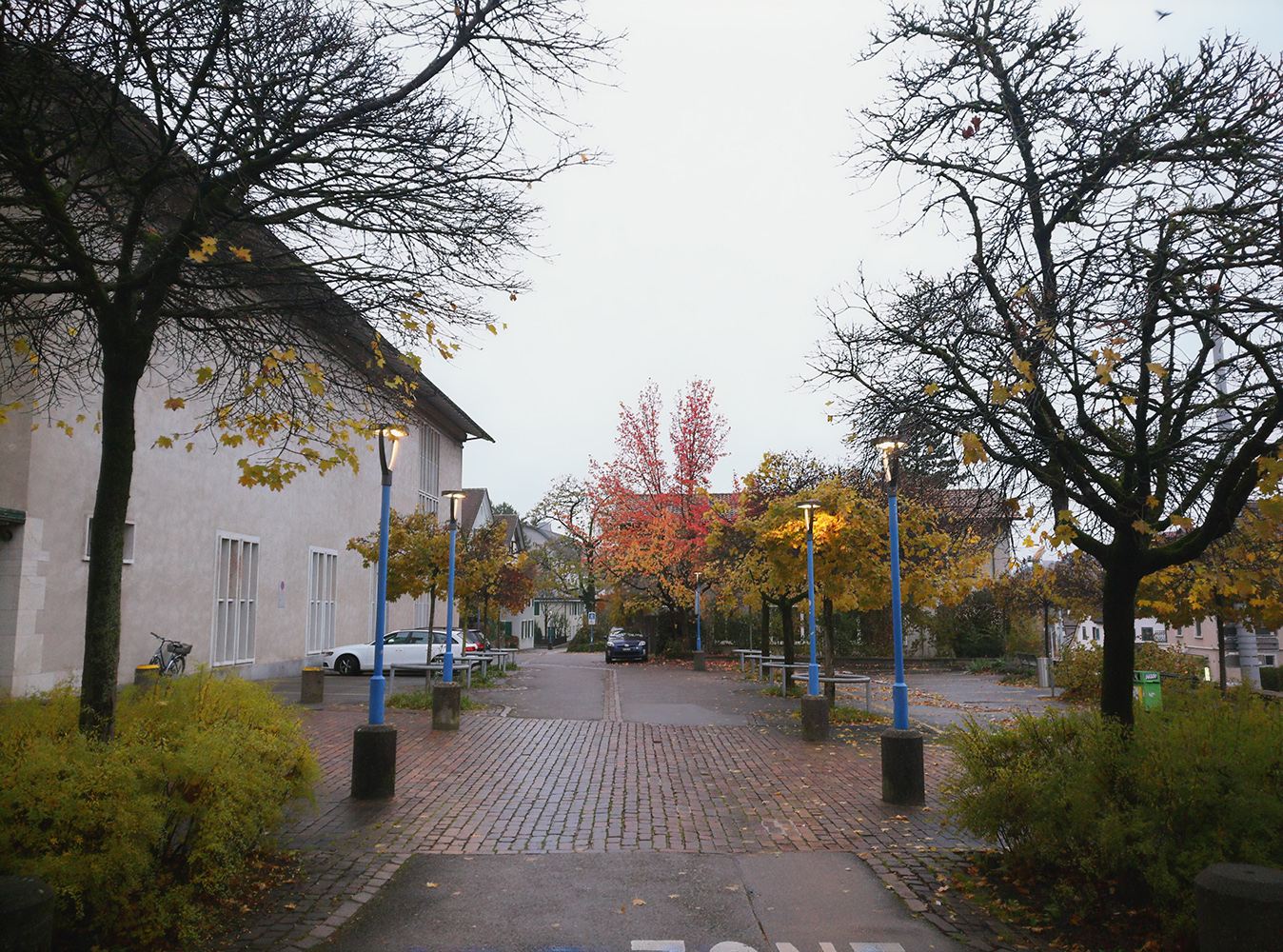}&
    \includegraphics[width=0.24\linewidth]{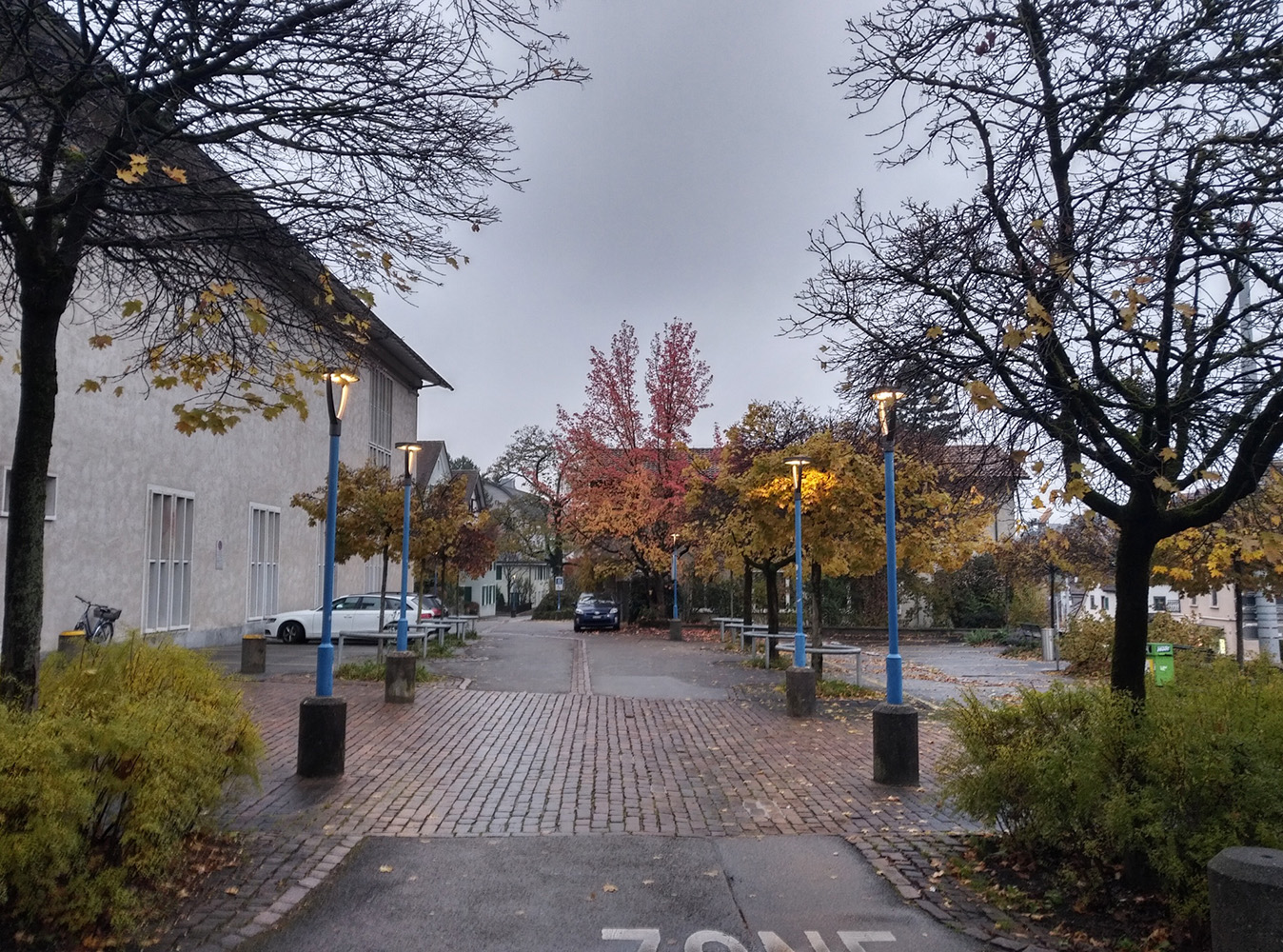}\\
    \includegraphics[width=0.24\linewidth]{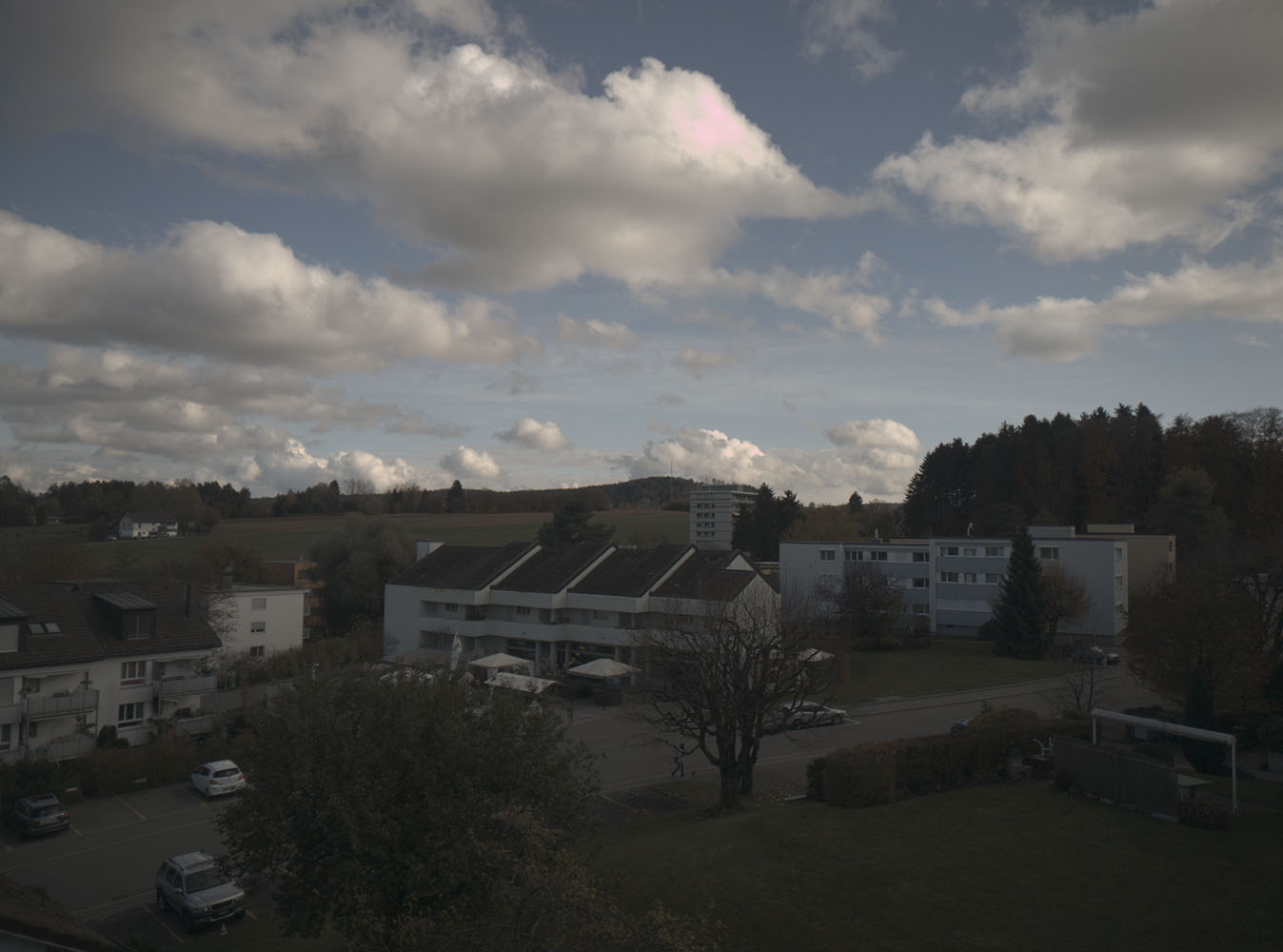}&
    \includegraphics[width=0.24\linewidth]{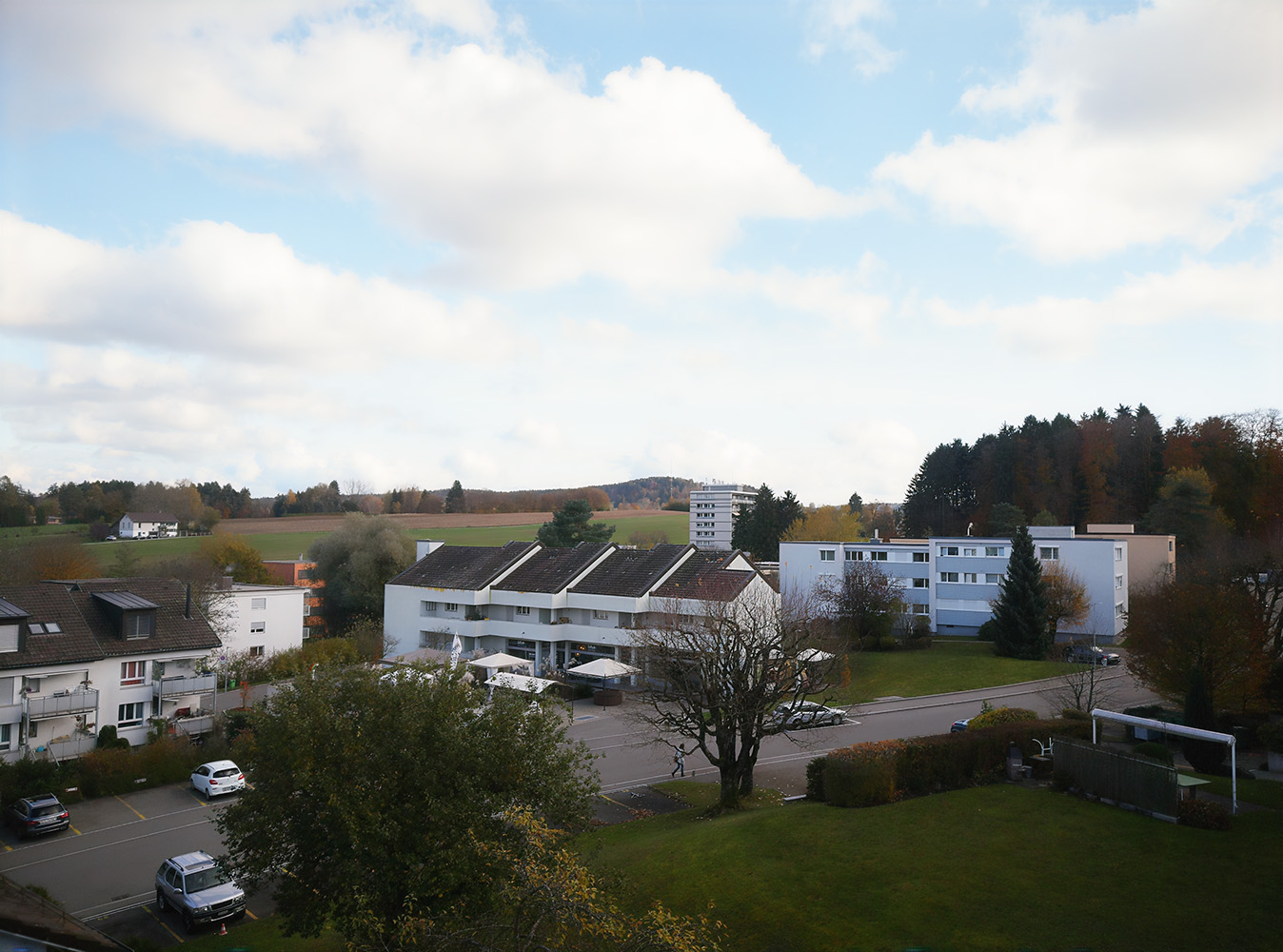}&
    \includegraphics[width=0.24\linewidth]{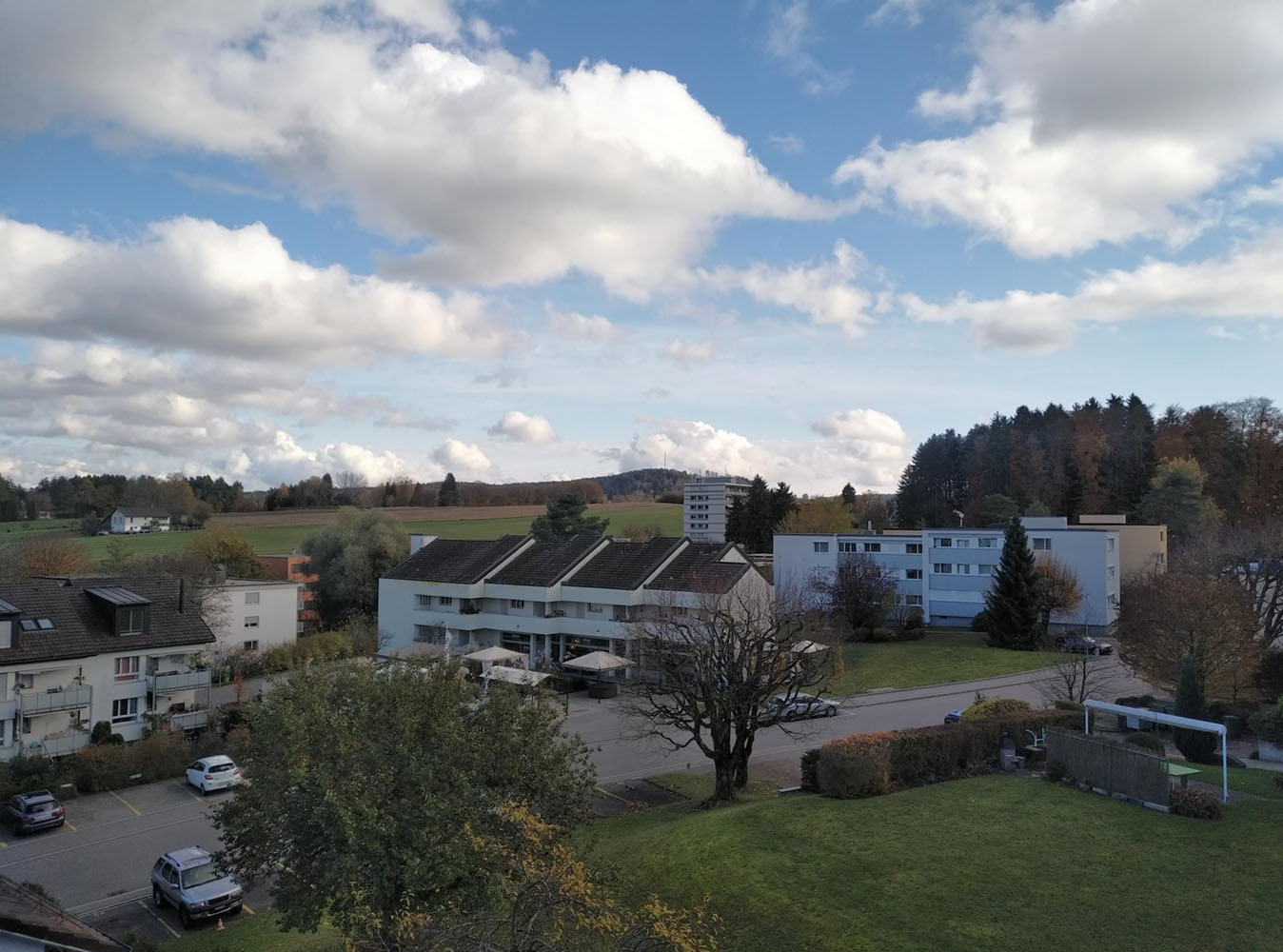}
\end{tabular}
}
\vspace{0.2cm}
\caption{\small{The results of the proposed method on RAW images from the BlackBerry KeyOne smartphone. From left to right: the original visualized RAW image, reconstructed RGB image and the same photo obtained with KeyOne's built-in ISP system using HDR mode.}}
\label{fig:blackberry}
\end{figure*}

\subsection{User study}
The ultimate goal of our work is to provide an alternative to the existing handcrafted ISPs and, starting from the camera's raw sensor readings, to produce DSLR-quality images for the end user of the smartphone. To measure the overall quality of our results, we designed a user study with the Amazon Mechanical Turk~\footnote{\url{https://www.mturk.com}} platform.

For the user study we randomly picked test raw input images in full resolution to be processed by 3 ISPs (the basic Visualized RAW, Huawei P20 ISP, and PyNET). The subjects were asked to assess the quality of the images produced by each ISP solution in direct comparison with the reference images produced by the Canon 5D Mark IV DSLR camera. The rating scale for the image quality is as follows: 1 - `much worse', 2 - `worse', 3 - `comparable', 4 - `better', and 5 - `much better' (image quality than the DSLR reference image). For each query comprised from an ISP result versus the corresponding DSLR image, we collected opinions from 20 different subjects. For statistical relevance we collected 5 thousand such opinions.


The Mean Opinion Scores (MOS) for each ISP approach are reported in Table~\ref{tab:MOS}. We note again that 3 is the MOS for image quality that is `comparable' to the DSLR camera, while 2 corresponds to a clearly `worse' quality. In this light, we conclude that the Visualized RAW ISP with a score of 2.01 is clearly `worse' than the DSLR camera, while the ISP of the Huawei P20 camera phone gets 2.56, almost half way in between the `worse' and `comparable'. Our PyNET ISP, on the other hand, with a score of 2.77 is substantially better than the innate ISP of the P20 camera phone, but also below the quality provided by the Canon 5D Mark IV DSLR camera.

\smallskip

\begin{table}[t!]
    \centering
    \begin{tabular}{c|c|c}
RAW input & ISP & MOS $\uparrow$ \\
    \hline
    &Visualized RAW & 2.01\\
Huawei P20 &Huawei P20 ISP & 2.56\\
    &{PyNET (ours)} & 2.77\\
    \hline
\multicolumn{2}{c|}{\textit{Canon 5D Mark IV}} & \textit{3.00}\\
    \end{tabular}
    \vspace{0mm}
    \caption{\small{Mean Opinion Scores (MOS) obtained in the user study for each ISP solution in comparison to the target DSLR camera (3~-- comparable image quality, 2~-- clearly worse quality).}}
    \label{tab:MOS}
    \vspace{-1.8mm}
\end{table}

In a direct comparison between the Huawei P20 ISP and our PyNET model (used now as a reference instead of the DSLR) with the same protocol and rating scale, we achieved a MOS of 2.92. This means that the P20's ISP produces images of poorer perceptual quality than our PyNET when starting from the same Huawei P20 raw images.

\subsection{Generalization to Other Camera Sensors}

\begin{figure*}[t!]
\centering
\setlength{\tabcolsep}{1pt}
\resizebox{\linewidth}{!}
{
\begin{tabular}{cccc}
\scriptsize{BlackBerry RAW Image (Visualized)}\normalsize & \scriptsize{Reconstructed RGB Image (PyNET)}\normalsize & \scriptsize{BlackBerry KeyOne ISP Image}\normalsize\\
    \includegraphics[width=0.33\linewidth]{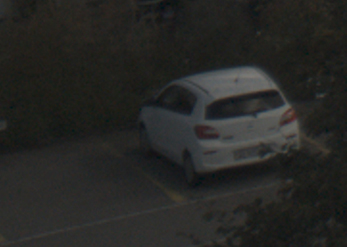}&
    \includegraphics[width=0.33\linewidth]{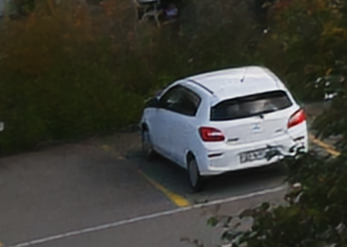}&
    \includegraphics[width=0.33\linewidth]{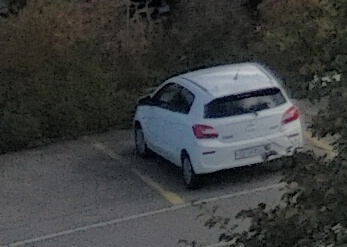}\\
    \includegraphics[width=0.33\linewidth]{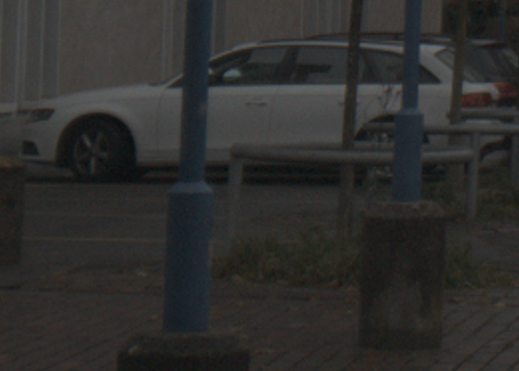}&
    \includegraphics[width=0.33\linewidth]{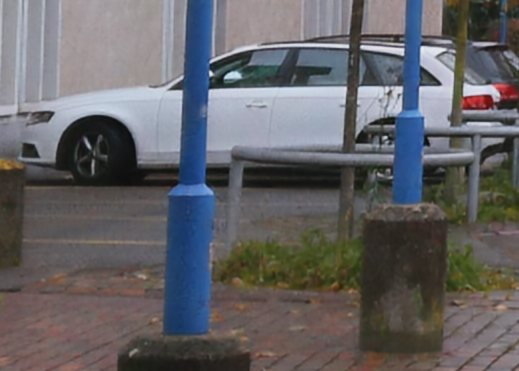}&
    \includegraphics[width=0.33\linewidth]{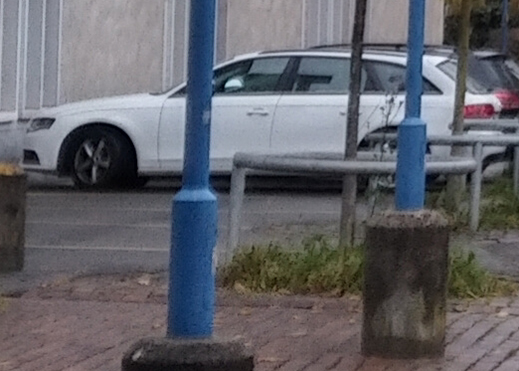}
\end{tabular}
}
\vspace{0.0cm}
\caption{\small{Image crops from the BlackBerry KeyOne RAW, reconstructed and ISP images, respectively.}}
\label{fig:blackberry_crops}
\vspace{-0.2cm}
\end{figure*}

While the proposed deep learning model was trained to map RAW images from a particular device model / camera sensor, we additionally tested it on a different smartphone to see if the learned manipulations can be transferred to other camera sensors and optics. For this, we have collected a number of images with the BlackBerry KeyOne smartphone that also has a 12 megapixel main camera, though is using a different sensor model (Sony IMX378) and a completely different optical system. RAW images were collected using the \textit{Snap Camera HDR}~\footnote{\tiny{\url{https://play.google.com/store/apps/details?id=com.marginz.snaptrial}}} Android application, and we additionally shoot the same scenes with KeyOne's default camera app taking photos in HDR mode. The obtained RAW images were then fed to our pre-trained PyNET model, the resulting reconstruction results are illustrated in Figure~\ref{fig:blackberry}.

As one can see, the PyNET model was able to reconstruct the image correctly and performed an accurate recovery of the colors, revealing many color shades not visible on the photos obtained with BlackBerry's ISP. While the latter images have a slightly higher level of details, PyNET has removed most of the noise present on the RAW photos as shown in Figure~\ref{fig:blackberry_crops} demonstrating smaller image crops. Though the reconstructed photos are not ideal in terms of the exposure and sharpness, we should emphasize that the model was not trained on this particular camera sensor module, therefore much better results can be expected when tuning PyNET on the corresponding RAW--RGB dataset.

\section{Conclusions}
\label{sec:conclusions}

In this paper, we have investigated and proposed a change of paradigm -- replacing an existing handcrafted ISP pipeline with a single deep learning model. For this, we first collected a large dataset of RAW images captured with the Huawei P20 camera phone and the corresponding paired RGB images from the Canon 5D Mark IV DSLR camera. Then, since the RAW to RGB mapping implies complex global and local non-linear transformations, we introduced PyNET, a versatile pyramidal CNN architecture. Next, we validated our PyNET model on the collected dataset and achieved significant quantitative PSNR and MS-SSIM improvements over the existing top CNN architectures. Finally, we conducted a user study to assess the perceptual quality of our ISP replacement approach. PyNET proved better perceptual quality than the handcrafted ISP innate to the P20 camera phone and closer quality to the target DSLR camera. We conclude that the results show the viability of our approach of an end-to-end single deep learned model as a replacement to the current handcrafted mobile camera ISPs. However, further study is required to fully grasp and emulate the flexibility of the current mobile ISP pipelines.

{\small
\bibliographystyle{ieee_fullname}

}

\end{document}